\documentclass[runningheads]{llncs}
\usepackage{graphicx}
\usepackage{subcaption}
\usepackage{times}
\usepackage{epsfig}
\usepackage{graphicx}
\usepackage{amsmath}
\usepackage{amssymb}
\usepackage{pifont}
\usepackage{multirow}
\usepackage{booktabs}
\usepackage{comment}
\usepackage{adjustbox}
\usepackage{my_ref}
\newcommand{\myparagraph}[1]{\vspace{4pt}\noindent\textbf{#1}}
\usepackage{hhline}
\usepackage{mathtools}

\usepackage{gensymb,siunitx} 
\usepackage{textcomp}
\usepackage[dvipsnames]{xcolor}

\newcommand{\ie}{\textit{i}.\textit{e}. }
\newcommand{\eg}{\textit{e}.\textit{g}. }
\newcommand{\wrt}{w.r.t }
\begin{document}
\title{Learning Semantics for Visual Place Recognition through Multi-Scale Attention}
%
%
\author{Valerio Paolicelli\inst{1} \and
Antonio Tavera\inst{1} \and 
Carlo Masone\inst{2} \and
Gabriele Berton\inst{1} \and
Barbara Caputo\inst{1}
}
%
\authorrunning{Paolicelli et al.}
%
\institute{Polytechnic University of Turin, Turin, Italy \and
CINI - Consorzio Interuniversitario Nazionale per l’Informatica, Rome, Italy
%
\email{\\\inst{1}\{name.surname}@polito.it\}}
\maketitle              
\begin{abstract}
In this paper we address the task of visual place recognition (VPR), where the goal is to retrieve the correct GPS coordinates of a given query image against a huge geotagged gallery. While recent works have shown that building descriptors incorporating semantic and appearance information is beneficial, current state-of-the-art methods opt for a top down definition of the significant semantic content. Here we present the first VPR algorithm that learns robust global embeddings from both visual appearance and semantic content of the data, with the segmentation process being dynamically guided by the recognition of places through a multi-scale attention module.
Experiments on various scenarios validate this new approach and demonstrate its performance against state-of-the-art methods.
Finally, we propose the first synthetic-world dataset suited for both place recognition and segmentation tasks.
\end{abstract}
\section{Introduction}
\noindent
Visual place recognition (VPR)~\cite{Masone_IEEEAccess2021}, \ie, the task of recognizing the location where a photo was taken, is usually cast as an image retrieval problem: the location of a photo (\emph{query}) is estimated by comparing it to a huge database of geo-tagged images (\emph{gallery}). Much of the recent research on this subject has focused on finding better image representations to perform the retrieval.
Despite the advances in this direction made possible by the use of deep convolutional neural networks (DCNN) \cite{Arandjelovic_CVPR2016,Radenovic_PAMI2019,Tolias_ICLR2016}, current VPR solutions still fail to achieve the degree of generality and flexibility required to work across different environments and conditions \cite{zaffar-2019}. 
Recent studies have found that these problems can be mitigated by building image descriptors based not only on visual appearance, but also on the semantic content in the scene \cite{Naseer_ICRA2017,Larsson_ICCV2019,Hanjiang_TIP2021}. Intuitively, dynamic objects or elements that are both common across all places and that lack distinctive features (\eg, roads and sky) are not very informative for VPR. On the other hand, content that is stable across different conditions and that has a wide range of intra-class variations, such as buildings \cite{Naseer_ICRA2017}, can more robustly describe places. 

\begin{figure}[t!]
    \centering
    \includegraphics[width=8cm]{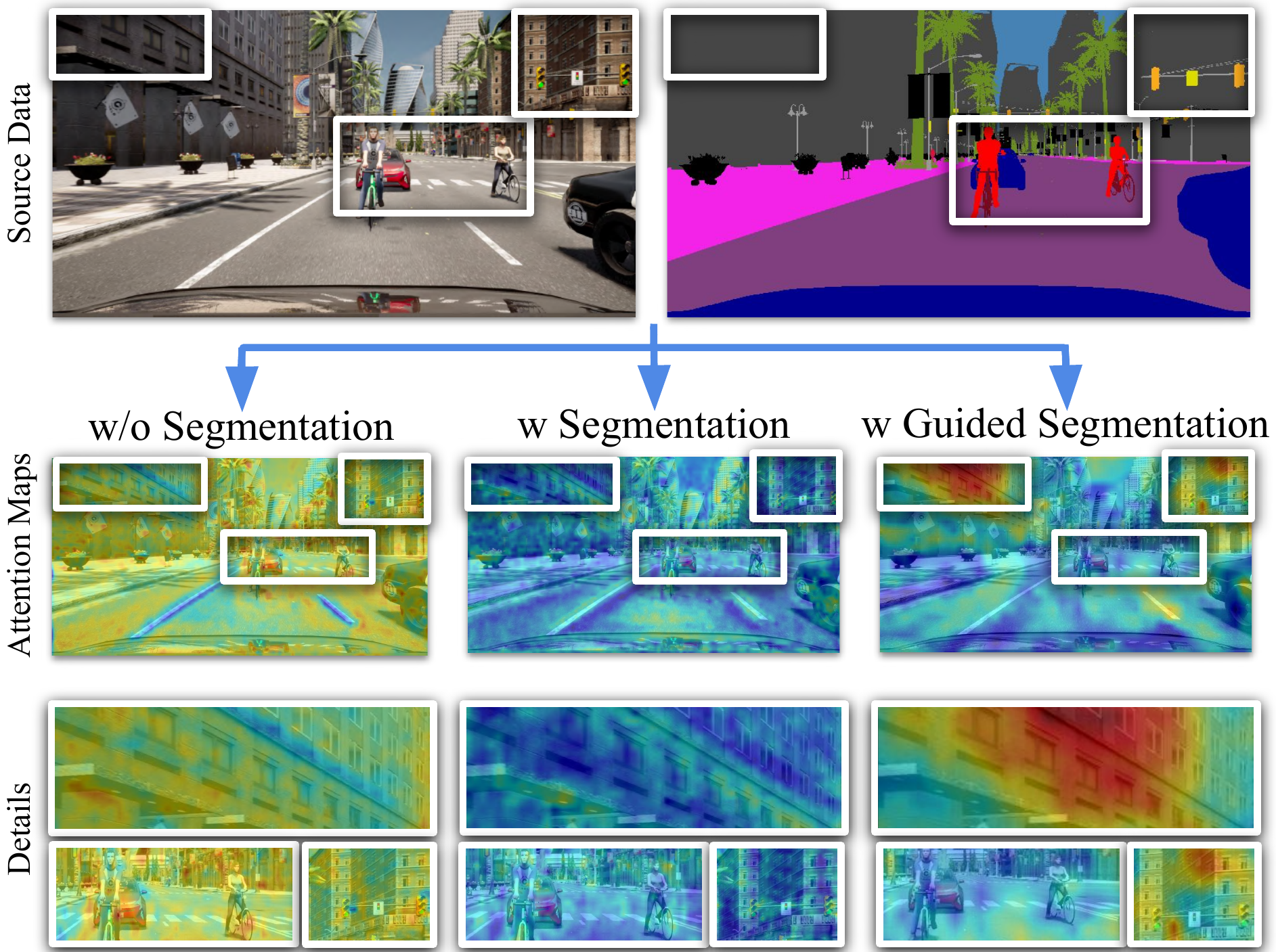}
    \caption{Prior works have shown that the VPR task can be improved by imbuing the features with semantic information. However, when the semantic task is not controlled by the place recognition, it draws information from all the scene indistinctly (middle). By conditioning the semantic segmentation on the place recognition task (right), the model learns to draw information only from the semantic categories that are most discriminative for a place, \eg, buildings and traffic lights.
    }
    \label{fig:teaser}
    \vspace{-10pt}
\end{figure}

While this intuition holds promise, it currently faces two intertwined open challenges, \ie, the need to define a priori the semantic content to use and the lack of an appropriate annotated database from where to learn it. Indeed, previous works on combining appearance and semantic information to generate global descriptors for VPR have either empirically defined what semantic classes should be used to describe places \cite{Mousavian-2015,Naseer_ICRA2017} or used all the semantic content available \cite{Hanjiang_TIP2021}. While this can be seen as a legitimate choice, it remains that it does not exist a public VPR dataset containing images annotated with pixel-wise semantic maps. 
What is currently available to the community are large-scale, multi-scenario VPR datasets without semantic annotations \cite{Maddern_IJRR2017,Torii_2015PAMI,Berton_WACV2021,Warburg_2020_CVPR}, or autonomous driving datasets that provide pixel-wise labels but either lack GPS annotation \cite{Cordts_CVPR2016,Ros2016Synthia,Richter2016GTA} or are too small in scale to effectively build query and gallery sets \cite{Yu_CVPR2020,Geiger_CVPR2012,Huang_arXiv2018,Geyer_ARXIV2020}.

Here we argue that it is best to let the model figure out what semantic information is more relevant to describe and recognize a place. Hence, we present a large scale synthetic database, annotated with both GPS and pixel-wise semantic maps, jointly with a new architecture that builds global descriptors for VPR in a data driven manner, informed by both visual appearance and semantic content of the training data. 
To do so, we introduce an attention-based mechanism to dynamically condition, during training, the segmentation process on the recognition of places.
The attention mechanism operates by leveraging features located at multiple spatial scales to capture the discriminative urban objects with different sizes in the scene.

In summary, the contributions of this paper are:
\begin{itemize}
    \item a new data-driven method to generate highly informative global descriptors for VPR, leveraging both visual appearance and semantic content at different scales;
    \item a new synthetic dataset for large-scale visual place recognition that also contains pixel-wise semantic labels;
    \item extensive validation on various real-world scenarios, demonstrating both the effectiveness of each component in our architecture and a consistent improvement over previous state-of-the-art methods.
\end{itemize}
\section{Related Work}
\myparagraph{Semantically informed Visual Place Recognition.}
Most modern approaches for visual place recognition rely on an image retrieval formulation \cite{Masone_IEEEAccess2021}, using DCNNs to extract appearance features which then generate global descriptors by means of aggregation \cite{Arandjelovic_CVPR2016} or pooling \cite{Tolias_ICLR2016,Radenovic_PAMI2019}.  
Recent works have proposed to enhance this paradigm by introducing attention mechanisms \cite{Xin_ICRA2019,Berton_WACV2021,Zhu_ACMMM2018,Kim_CVPR2017} and domain adaptation techniques to align features of different scenarios \cite{Berton_WACV2021,Jenicek_ICCV2019}. 
Few studies suggest building the global descriptors not only using the visual information in the images, but also their semantic content. 
Along these lines, the method presented in \cite{Naseer_ICRA2017} requires segmenting an images also at inference time 
while \cite{Larsson_ICCV2019} requires a 3D point cloud of the scene.
Closely related to our work is DASGIL \cite{Hanjiang_TIP2021}, an architecture that uses a single encoder shared by three tasks (VPR, depth mask reconstruction and semantic mask reconstruction) to create embeddings that fuse visual, geometric and semantic information. Similar to our solution, DASGIL is trained on a synthetic dataset and it uses domain adaptation to align the features extracted from the synthetic and real-world domains. 
Besides that, in \cite{Hanjiang_TIP2021} the segmentation task focuses indiscriminately on all the semantic classes. On the contrary, our solution is built on the intuition that not all the semantic content is useful for VPR and we let the place recognition task guide the segmentation one via an attention mechanism. Moreover, DASGIL builds a global descriptor by flattening and concatenating the features extracted at multi-scale, without an embedding step. This produces extremely large descriptors which are not well suited for large scale problems. Instead, we use a novel multi-scale aggregation layer which produces more compact descriptors.

\section{Method}\label{sec:method} 
\begin{figure*}[!t]
    \centering
    \includegraphics[width=\linewidth]{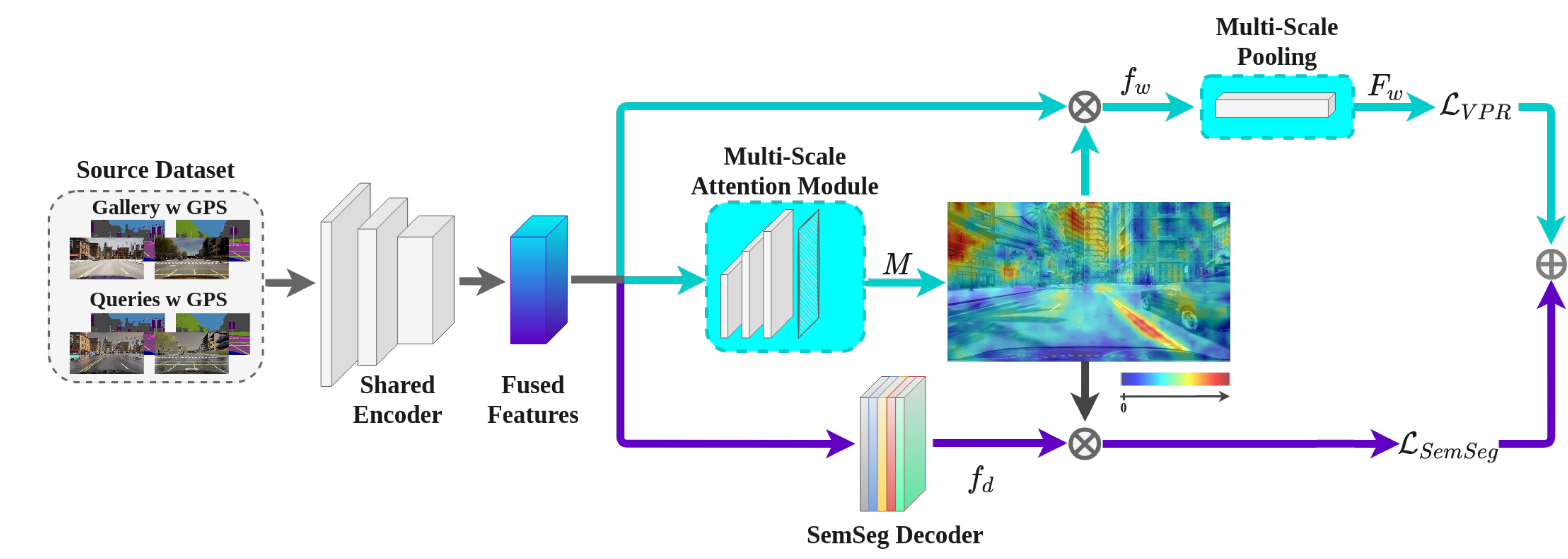}
    \caption{Illustration of the proposed architecture. The Multi-Scale attention module (middle) is trained by the place recognition task (top) and guides the segmentation module (bottom) towards relevant semantic classes. \textbf{\textcolor{cyan}{Blue lines}} refer to the visual place recognition task (VPR) that implements the novel multi-scale pooling layer (top). \textbf{\textcolor{violet}{Purple lines}} refer to the segmentation task (bottom).}
    \label{fig:architecture}
    \vspace{-6pt}
\end{figure*}

\noindent 
We consider having at training a collection $X=\{(x, y, z)\}$ of $N$ triplets, where $x$ is an RGB image composed by $\mathcal{I}$ pixels, $y$ is the semantic map that associates to each pixel $i$ a class from a set of semantic classes $C$, and $z$ is the GPS coordinate where the image was taken.
We propose a novel framework for visual place recognition that leverages both the appearance and pixel-wise semantic information available during training to learn image representations that are more effective for the place recognition problem.
Our architecture, depicted in \cref{fig:architecture}, consists of a single encoder shared by two tasks: 
\begin{itemize}
    \item a visual place recognition task (\emph{VPR}) (\cref{sec:vpr_branch}), that implements a novel multi-scale pooling layer to generate the global embeddings used for the retrieval process;
    \item a semantic segmentation task (\emph{SemSeg}) (\cref{sec:semseg_branch}), that implements a decoder for parsing the scene according to the set of classes $C$.
\end{itemize}
By sharing the same encoder, the two tasks induce it to learn features that combine both the visual information used for VPR and the semantic information present in the scene. However, without a proper mechanism to control this fusion, the model would equally focus on all semantic classes, regardless of their actual relevance for the VPR task. 
We introduce such a mechanism in the form of a multi-scale attention module that is trained only on the VPR task but modulates also the features extracted by the SemSeg decoder. In this way, the VPR guides the semantic segmentation, informing it where to focus on the scene.
Note that at inference time the SemSeg decoder is not used, thus the deployed model is quite lightweight.

\subsection{Multi-scale and attentive VPR task} 
\label{sec:vpr_branch}
\noindent
The shared encoder in our architecture is a ResNet \cite{He_CVPR2016} truncated after the last convolutional block. We indicate as $f_4$ and $f_5$ the outputs of the last two convolutional blocks, \emph{conv4} and \emph{conv5}, with shapes $C_4 \times H_4 \times W_4$ and $C_5 \times H_5 \times W_5$, respectively.
These features are used as input to both the multi-scale attention module (\cref{fig:architecture}, middle)
and to the novel multi-scale pooling layer (\cref{fig:architecture}, top).

\begin{figure}[!tbp]
  \centering
  \begin{minipage}[b]{0.45\textwidth}
    \includegraphics[width=\textwidth]{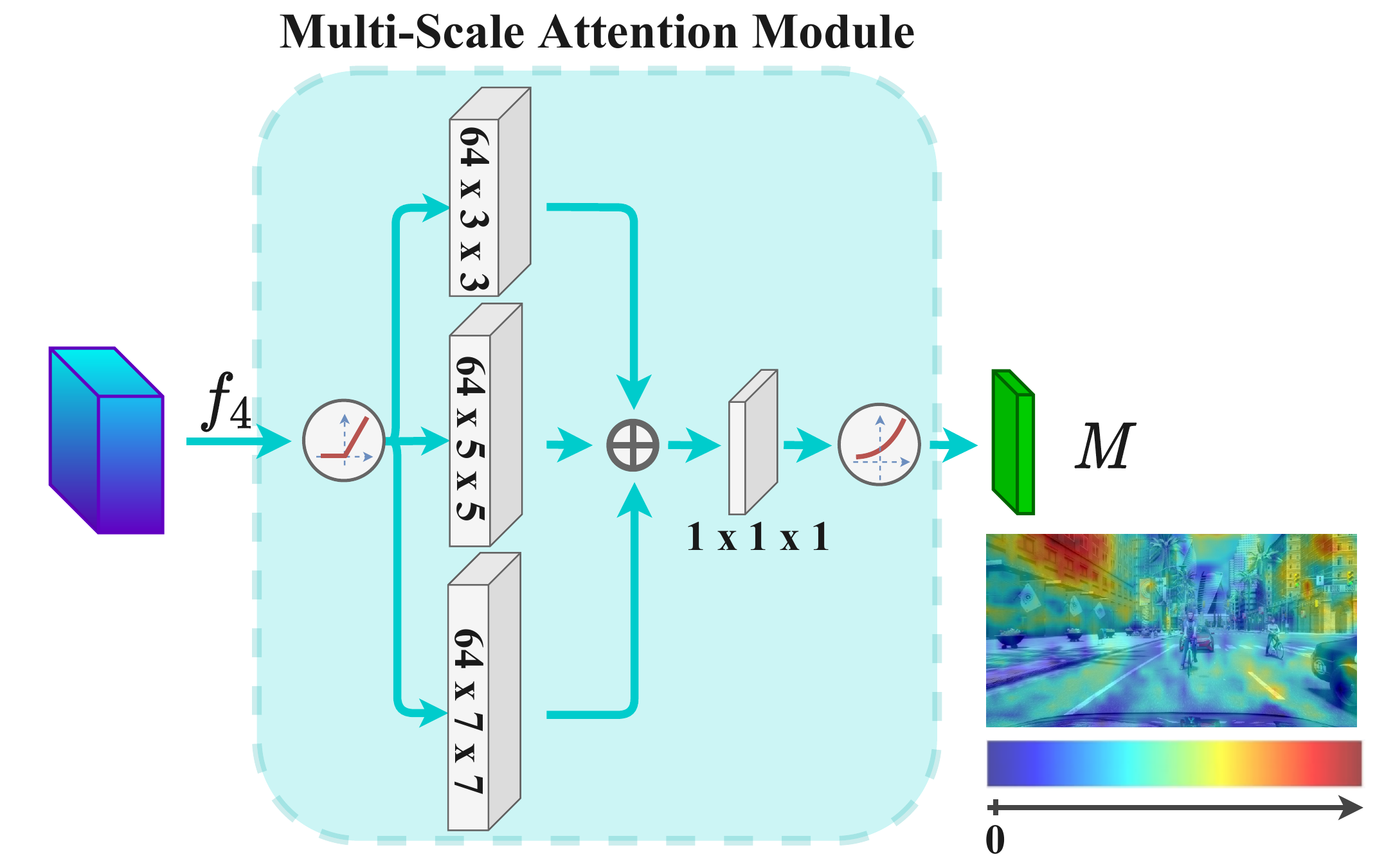}
    \caption{The attention module leverages features at multiple spatial scales to capture objects with different sizes, and produces a map $M$ that marks the retrieval salient regions. $\oplus$ indicates up-sampling and concatenation.}
    \label{fig:attention}
  \end{minipage}
  \hfill
  \begin{minipage}[b]{0.53\textwidth}
    \includegraphics[width=\linewidth]{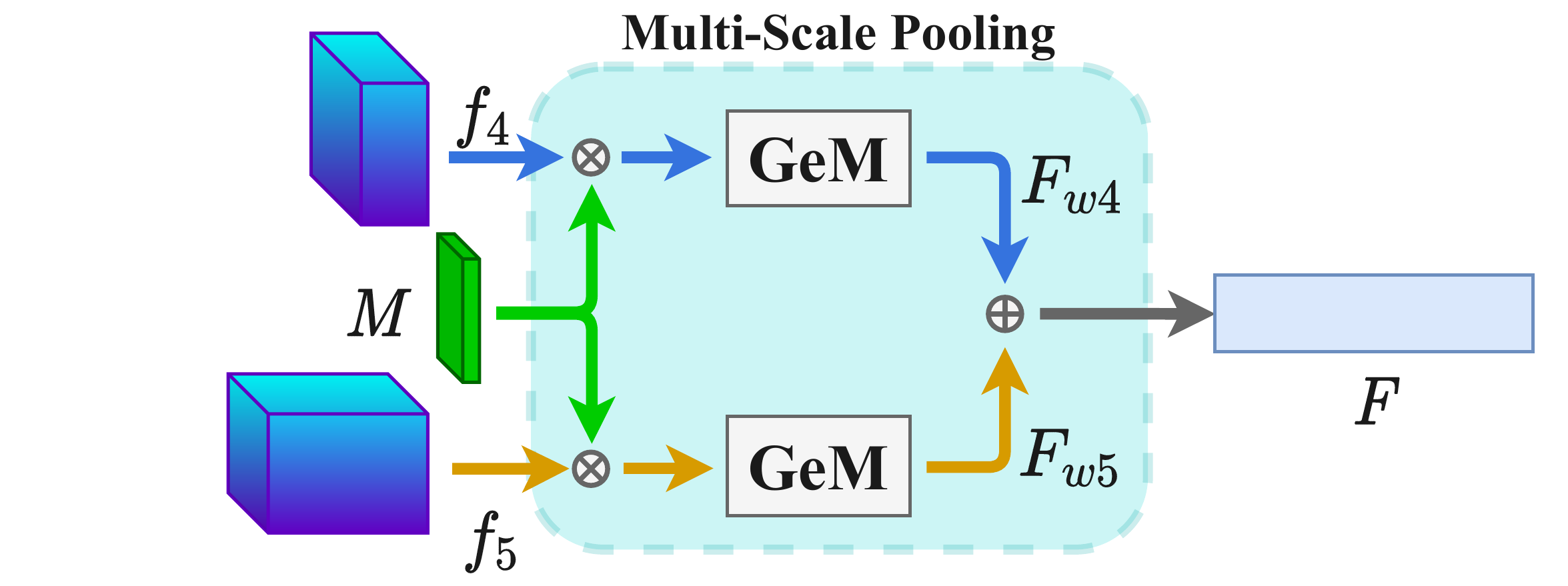}
    \caption{Illustration of the multi-scale pooling module. Features extracted at different levels of the shared encoder are exploited to compute the global descriptors. $\otimes$ indicates up-sampling and dot product. $\oplus$ indicates concatenation.}
    \label{fig:pooling}
  \end{minipage}
\end{figure}

\myparagraph{Multi-scale attention.}
Recent works have demonstrated the use of multi-scale attention mechanisms in place recognition as a way to focus on the salient regions in the image \cite{Xin_ICRA2019,Zhu_ACMMM2018,Kim_CVPR2017}.
In our architecture, the attention module becomes instrumental to make place recognition guide the semantic segmentation during training.
The module, depicted in \cref{fig:attention}, takes the output $f_4$ from the encoder and passes it through a bank of $O \times k_s \times k_s$ filters with $O$ number of output channels and different $k_s$ kernel sizes ($64 \times 3 \times 3$, $64 \times 5 \times 5$ and $64 \times 7 \times 7$).
The outputs of these filters are upsampled and concatenated channel-wise, before passing through a $1 \times 1 \times 1$ filter and a softplus function that produces a $1 \times H \times W$ attention map $M$.
The scores in the attention map $M$ indicate where the retrieval is focusing.


\myparagraph{Multi-scale pooling.}
In VPR it is widely common to use pooling layers after the convolutional backbone to extract compact global descriptors for the retrieval, the state-of-the-art being GeM  \cite{Radenovic_PAMI2019}. In order to exploit semantic and appearance information at different abstraction levels, we introduce a multi-scale GeM layer (\emph{ms-GeM}) that uses both $f_4$ and $f_5$, as illustrated in \cref{fig:pooling}. These features are first weighted by the attention scores \textit{M} via dot-product and L2-normalized. Then, they are pooled using vanilla GeM layers to produce the global descriptors $F_{w4} \in \mathbb{R}^{C_4}$ and $F_{w5} \in \mathbb{R}^{C_5}$. Finally, these descriptors are concatenated to form a representation $F \in \mathbb{R}^{C_4 + C_5}$.

\myparagraph{VPR loss.}
We use the weakly supervised triplet margin loss and training protocol from \cite{Arandjelovic_CVPR2016} to train the model to extract descriptors for the VPR task. For each training query we consider a positive and a negative example drawn from the gallery. The positive example is an image of the same place as the one depicted in the query, whereas the negative example is an image of a different location. Both the query and its corresponding positive/negative examples are from the source domain, therefore we use the available GPS information to select the examples. In particular, we consider as negative example the most similar image in the features space far from the query GPS coordinates.
Finally, for each query descriptor $F_q$ the loss is
\begin{equation}
    \mathcal{L}_{VPR} = h(d(F_{q}, F_{p}) + m - d(F_{q}, F_{n}))
    \label{eq:loss_vpr}
\end{equation}
where $h$ is the hinge loss, $d(\cdot,\cdot)$ is the Euclidean distance, $m$ is a fixed margin, $F_{p}$ and $F_{n}$ are the descriptors of the positive and negative examples, respectively. The goal pursued by $\mathcal{L}_{VPR}$ is to learn descriptors so that the distance between a training query and its positive example is smaller than the distance between the query and its negative example by at least the margin $m$.
Additionally, it is also used to optimize the parameters of the multi-scale attention module and thus focus the segmentation task on the salient regions for place recognition.


\subsection{Guided semantic segmentation task} \label{sec:semseg_branch}
\begin{figure}[t!]
    \centering
    \begin{minipage}{.24\linewidth}
        \begin{subfigure}{\linewidth}
            \includegraphics[width=\linewidth]{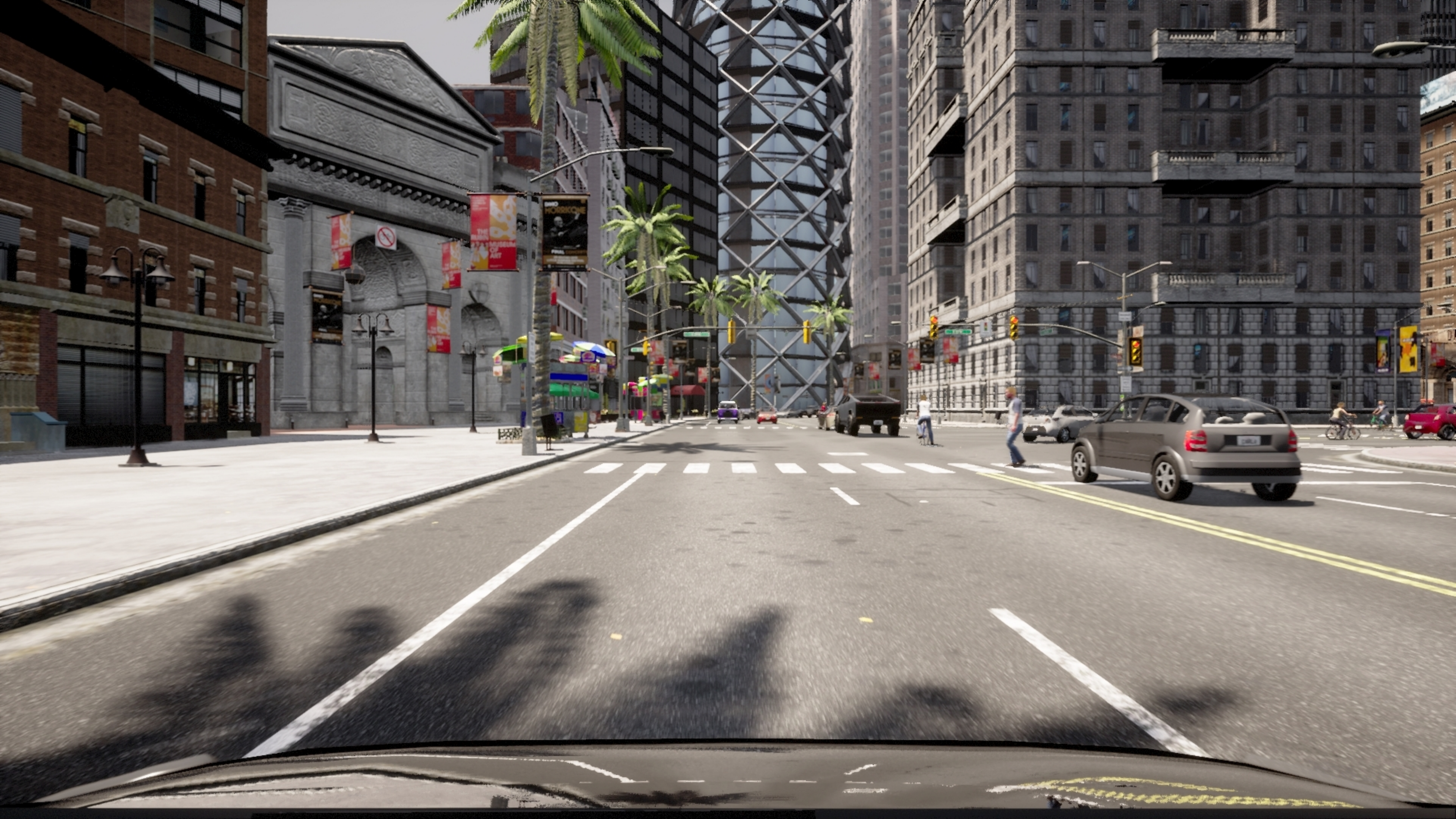} 
        \end{subfigure}
        \begin{subfigure}{\linewidth}
            \includegraphics[width=\linewidth]{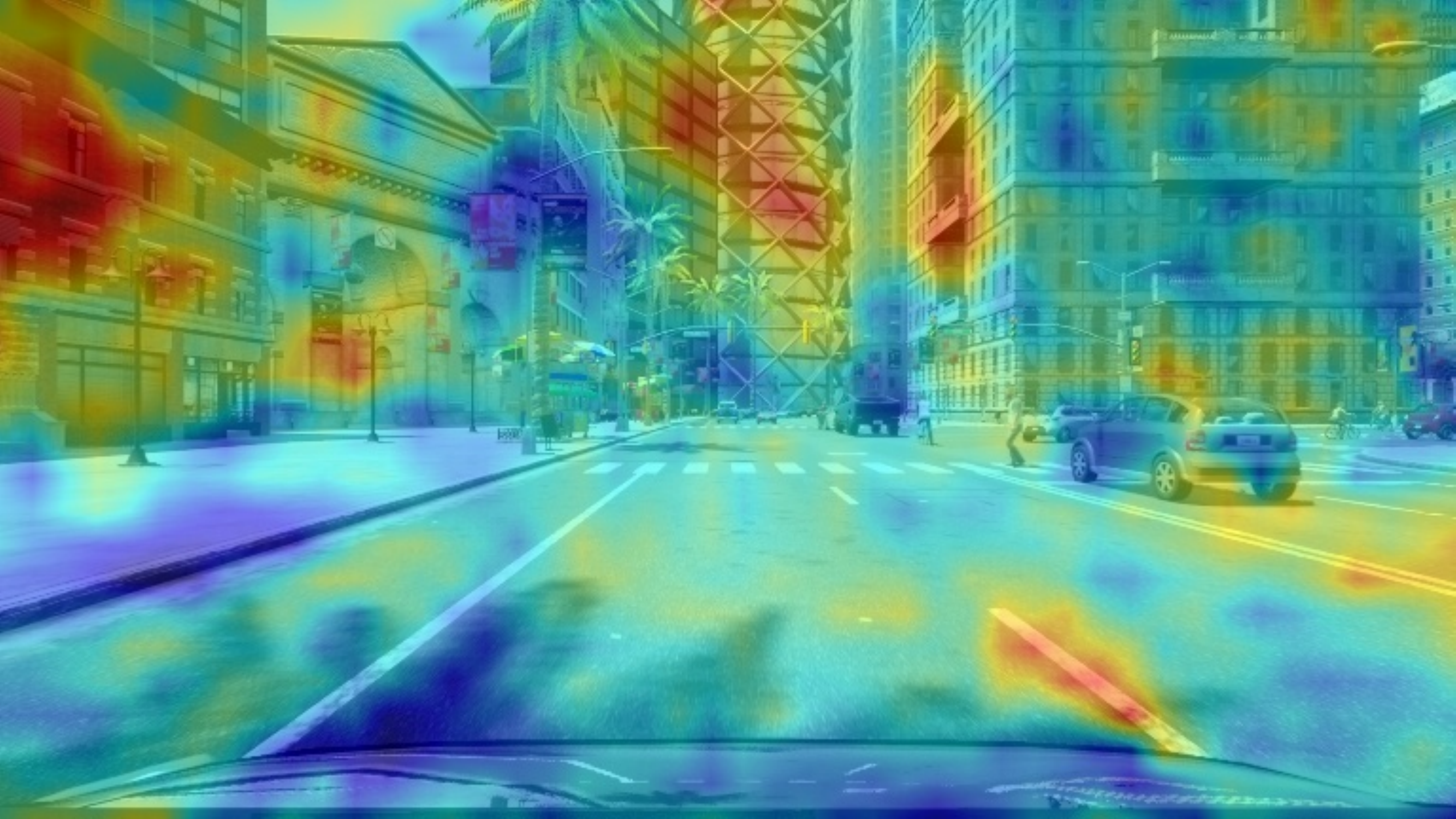}
        \end{subfigure}
        \label{fig:heatmaps_front}
    \end{minipage}
    \begin{minipage}{.24\linewidth}
        \begin{subfigure}{\linewidth}
            \includegraphics[width=\linewidth]{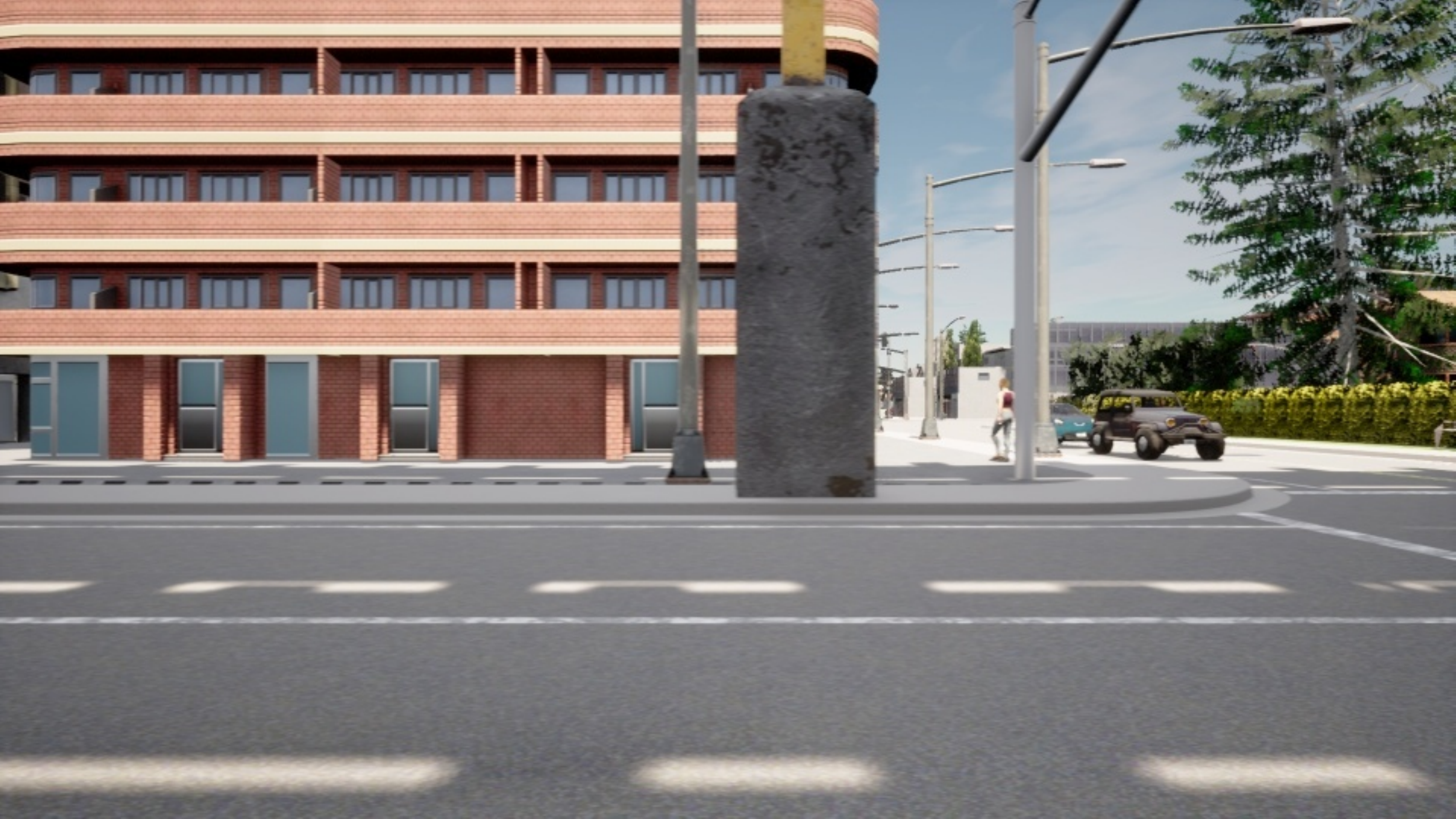} 
        \end{subfigure}
        \begin{subfigure}{\linewidth}
            \includegraphics[width=\linewidth]{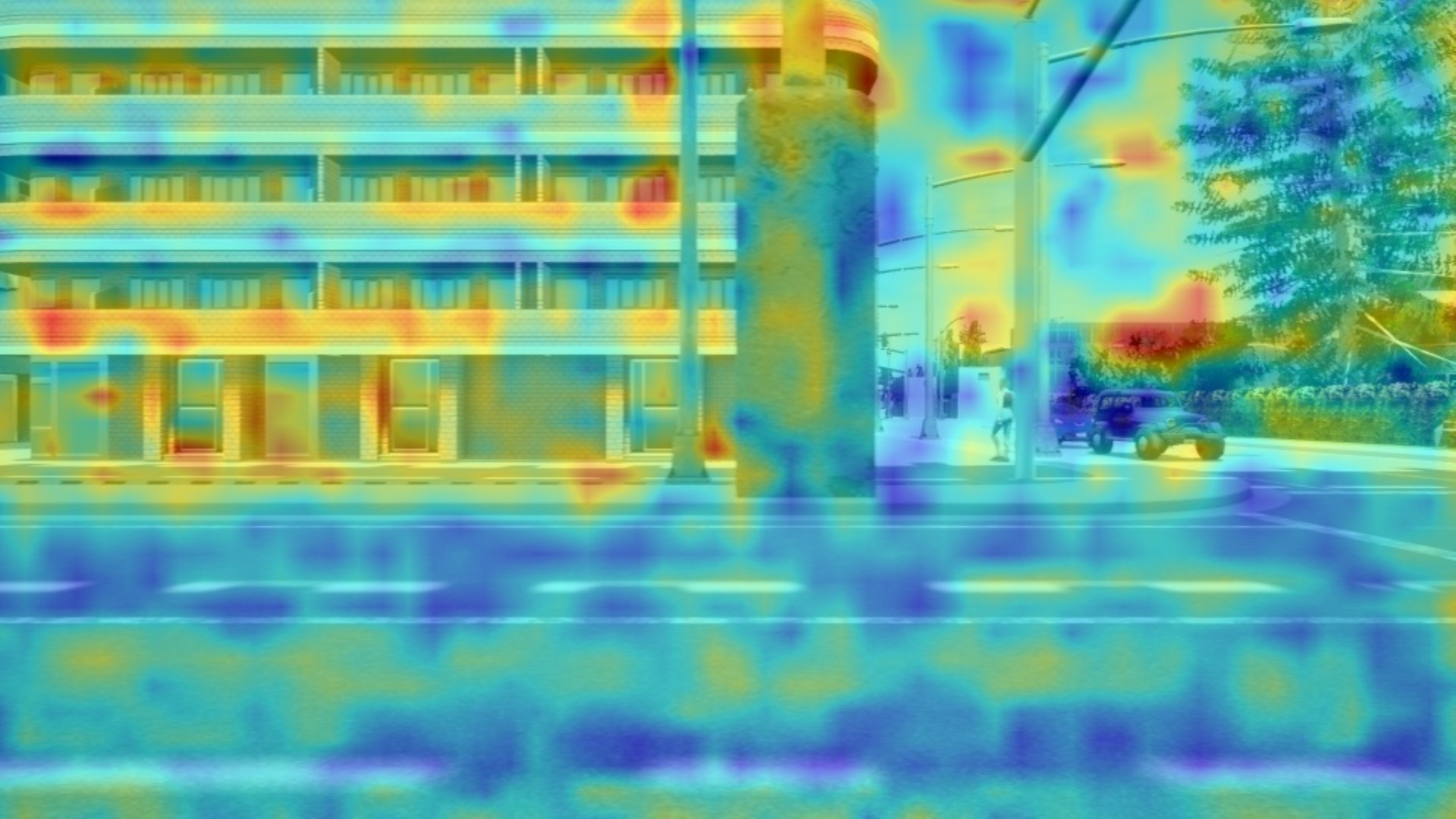}
        \end{subfigure}
        \label{fig:heatmaps_side}
    \end{minipage}
    \begin{minipage}{.24\linewidth}
        \begin{subfigure}{\linewidth}
            \includegraphics[width=\linewidth]{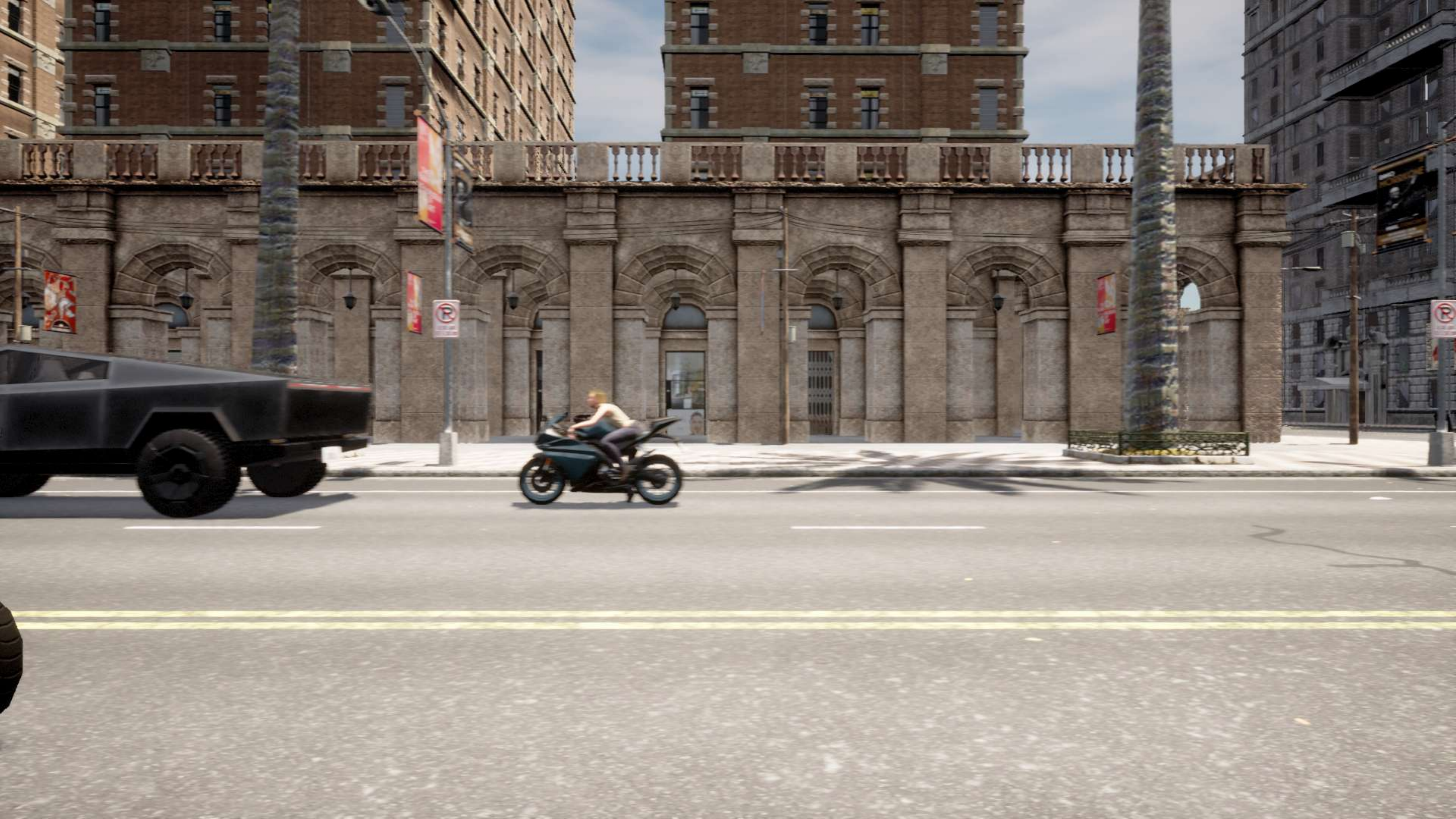} 
        \end{subfigure}
        \begin{subfigure}{\linewidth}
            \includegraphics[width=\linewidth]{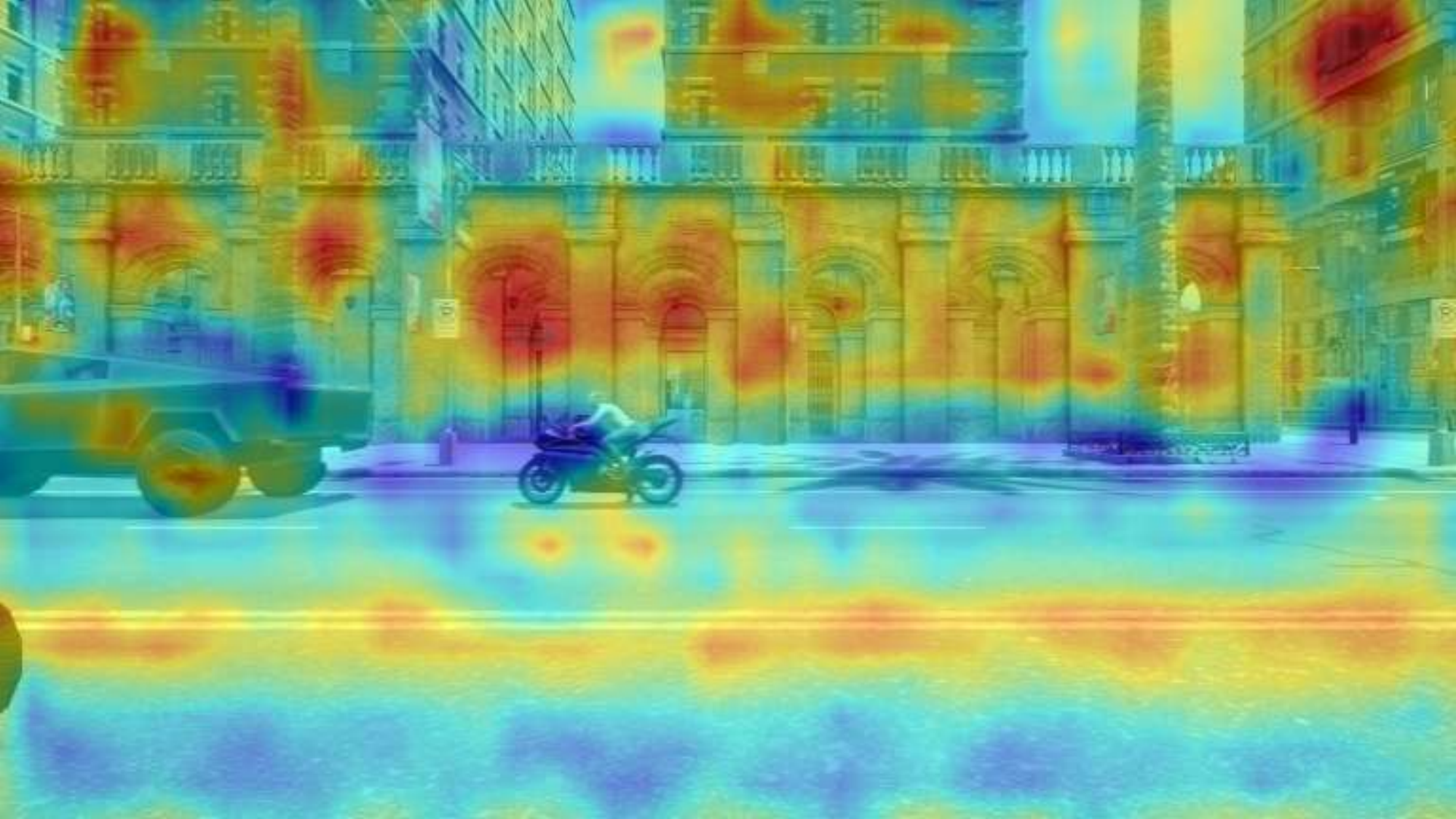}
        \end{subfigure}
        \label{fig:heatmaps_side1}
    \end{minipage}
    \begin{minipage}{.24\linewidth}
        \begin{subfigure}{\linewidth}
            \includegraphics[width=\linewidth]{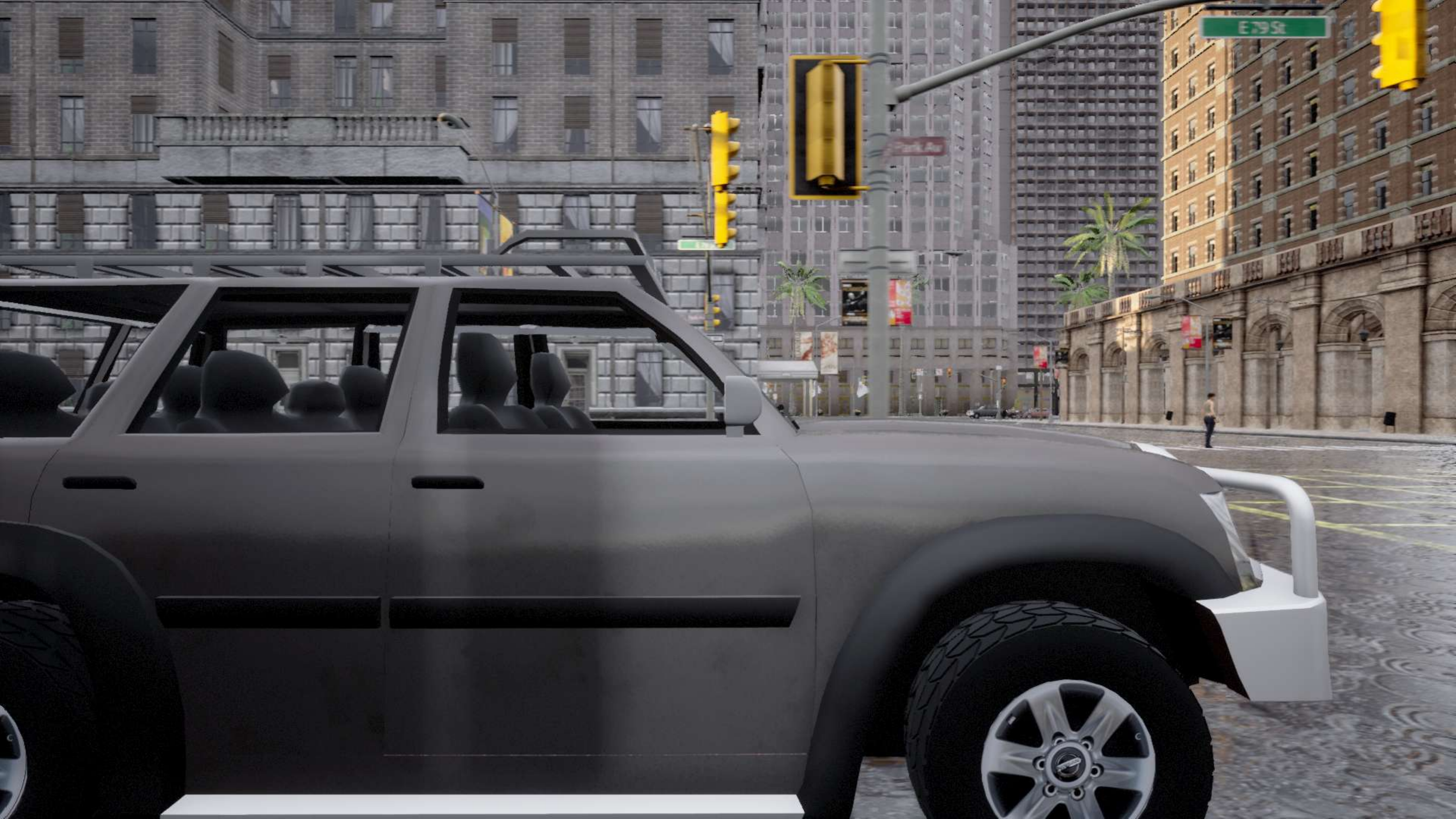} 
        \end{subfigure}
        \begin{subfigure}{\linewidth}
            \includegraphics[width=\linewidth]{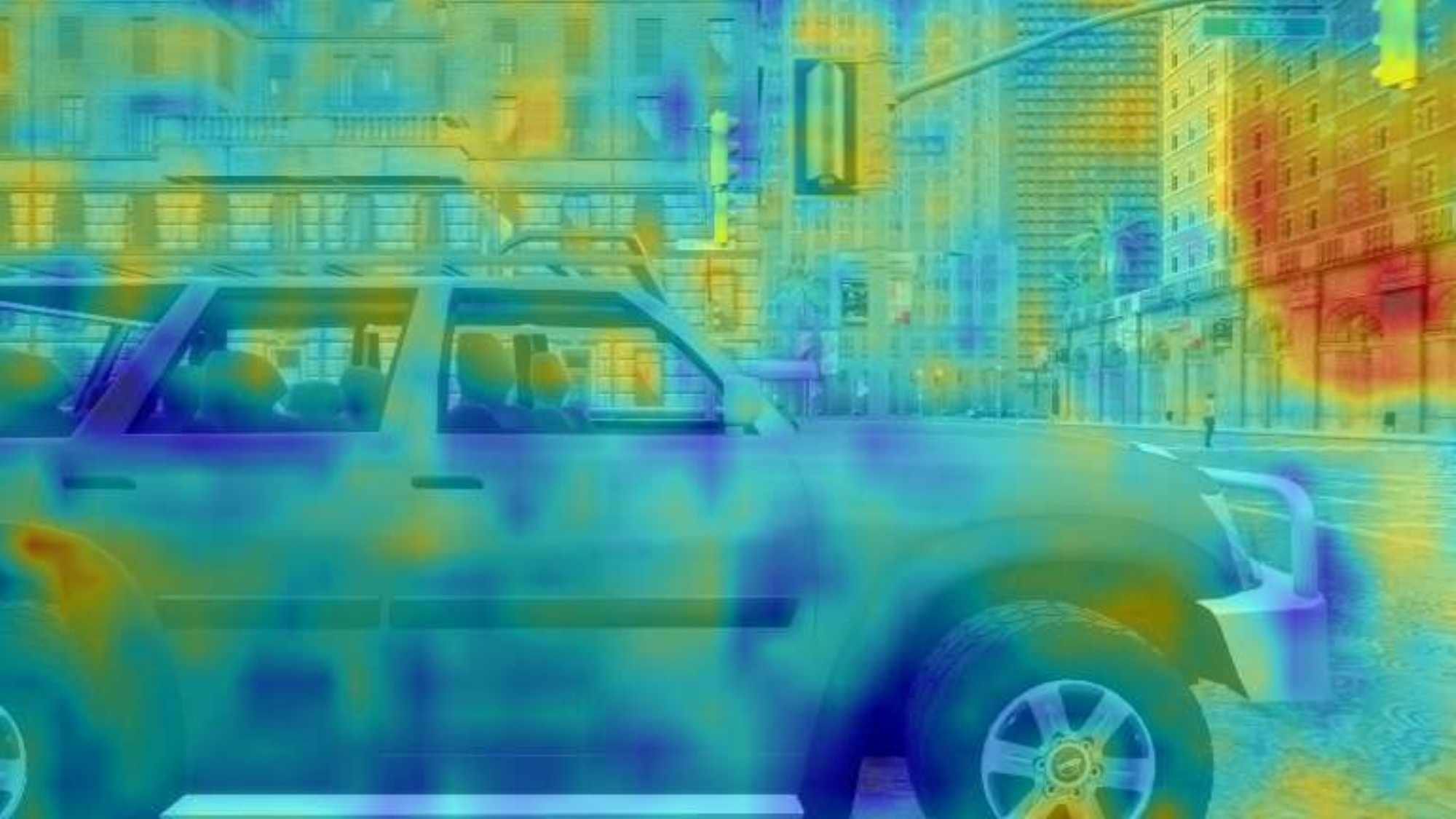}
        \end{subfigure}
        \label{fig:heatmaps_side2}
    \end{minipage}
    
    \vspace{5pt}
    \caption{Attention heatmaps extracted from our framework. We see that attention is focused only on relevant semantic classes according to the VPR task, such as buildings.}
    \label{fig:heatmaps}
    \vspace{-6pt}
\end{figure}

\noindent
The SemSeg task informs the features extraction process with semantic information. For this purpose, we use a semantic segmentation decoder $D_{seg}$ (see \cref{fig:architecture}).
However, to force the model to focus on the semantic information that is most discriminative for places, the output of the decoder $D_{seg}$ is weighted by the attention map $M$, which is trained by the VPR task alone.
To train the shared encoder and the decoder, the SemSeg branch uses a cross-entropy loss computed for each class $y_i$ at pixel $\mathcal{I}$,
i.e.,
\begin{equation}
\begin{aligned}
    \mathcal{L}_{SemSeg} = - \frac{1}{|\mathcal{I}|} \sum_{i \in \mathcal{I}} y_i \cdot \log p_i^{y_i} ((M^i \cdot f_{d}^i))
\end{aligned}
\label{eq:loss_semseg}
\end{equation}
where $M^i$ is the attention map related to the feature $f_{d}^i$ extracted from the decoder $D_{seg}$, while $p_i^{y_i}$ denotes the probability for class $y_i$ at pixel $\mathcal{I}$. \cref{fig:heatmaps} shows some examples of heatmaps resulting from the attention module on the source domain images. From these heatmaps it emerges that the network learns to focus on distinctive man-made structures such as buildings, shop signs and streetlights.

\subsection{Training loss}
\noindent
Summarizing, the VPR-SemSeg loss function is:
\begin{equation}
    \mathcal{L}_{VPR-SemSeg} = \mathcal{L}_{VPR} + \alpha \cdot \mathcal{L}_{SemSeg}
    \label{eq:loss_vpre_semseg}
\end{equation}
where $\alpha$ is a scalar weight.
Both $\mathcal{L}_{VPR}$ and $\mathcal{L}_{SemSeg}$ affect the encoder weights to produce features that are informative for place recognition and combine visual and semantic information.
However, only $\mathcal{L}_{VPR}$ impacts the attention module weights, thus making the segmentation dependent on the place recognition.

\section{A VPR and semantic segmentation dataset}

\noindent
The proposition of using semantic information in VPR in a data-driven approach is limited by the lack of a dataset that is both built for place recognition, \ie, containing multiple views of the same places tagged with GPS coordinates, and that provides fine-grained semantic labels.
There is only a couple of synthetic datasets that come close to this requirement, but eventually fall short. One is Virtual KITTI 2 \cite{cabon2020virtual}, which was used by \cite{Hanjiang_TIP2021}, but it only guarantees 447 images per scenario.  Moreover, images are not associated to GPS coordinates, so place matching is only done by the name of the images. 
The other is SYNTHIA \cite{Ros2016Synthia}, which contains very dense sequences of images that are suitable for the visual localization task~\cite{Pion-2020} ($\leq5$m from one gps coordinate to the other) but not for the coarser place recognition task usually considered in literature ($\leq25$m).
To develop our data-driven approach for VPR that combines both visual and semantic knowledge, and also to enable further research in this direction, we created a new synthetic dataset.

\begin{figure*}[t!]
    \centering
    \begin{minipage}{.24\linewidth}
        \begin{subfigure}{\linewidth}
            \includegraphics[width=\linewidth]{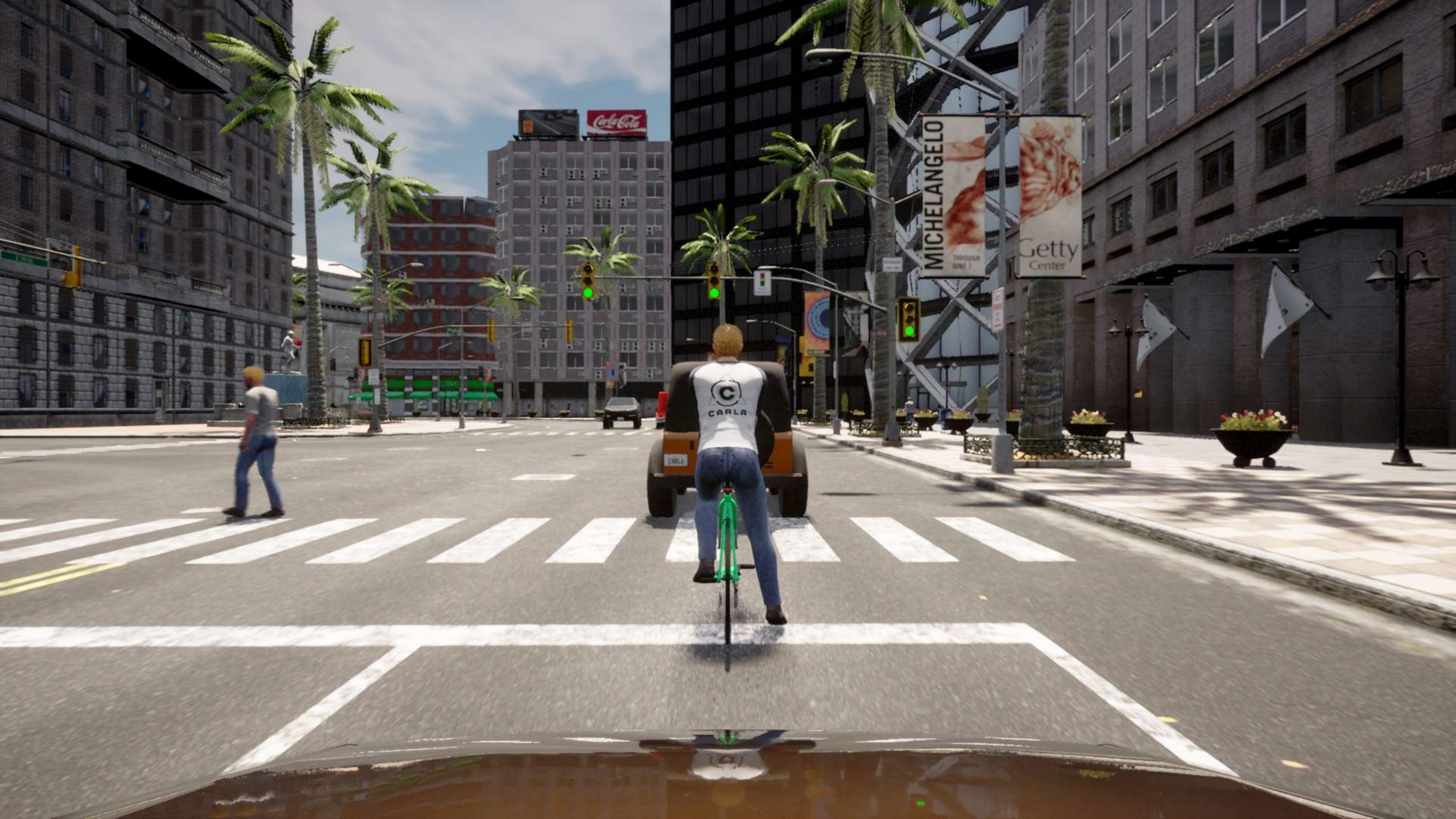}
        \end{subfigure}
        \begin{subfigure}{\linewidth}
            \includegraphics[width=\linewidth]{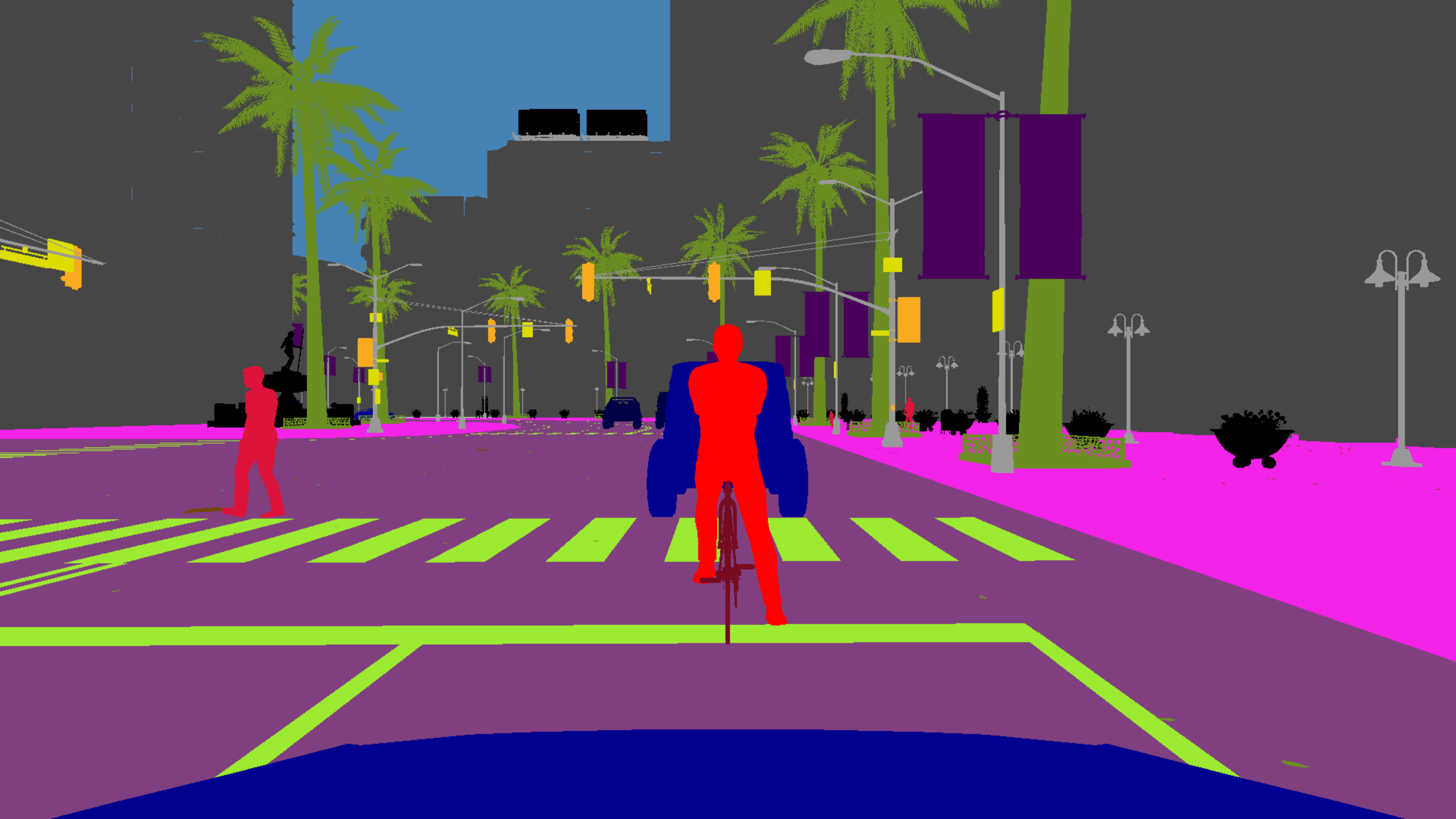}
        \end{subfigure}
        \subcaption{}
    \end{minipage}
    \begin{minipage}{.24\linewidth}
        \begin{subfigure}{\linewidth}
            \includegraphics[width=\linewidth]{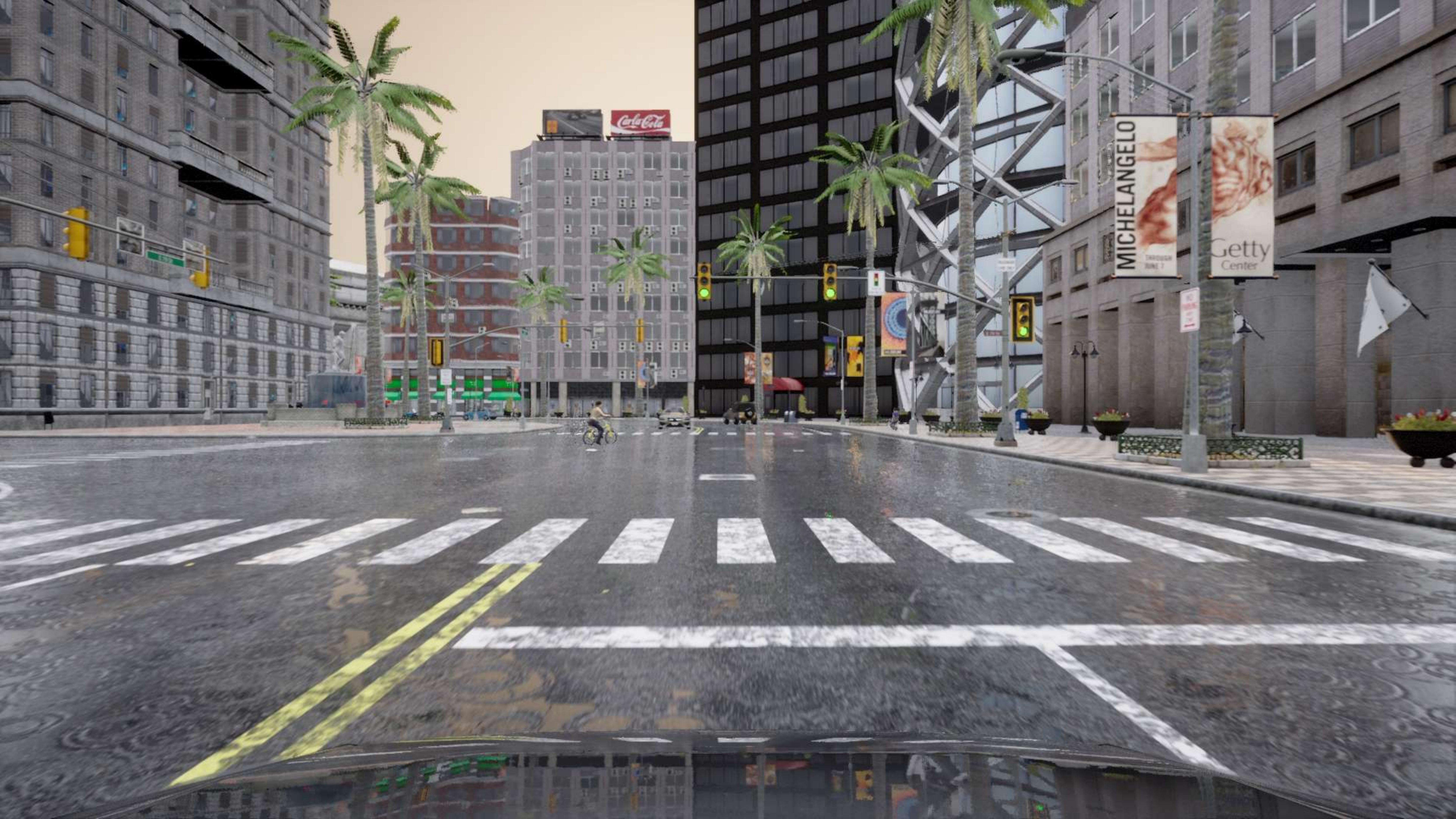}
        \end{subfigure}
        \begin{subfigure}{\linewidth}
            \includegraphics[width=\linewidth]{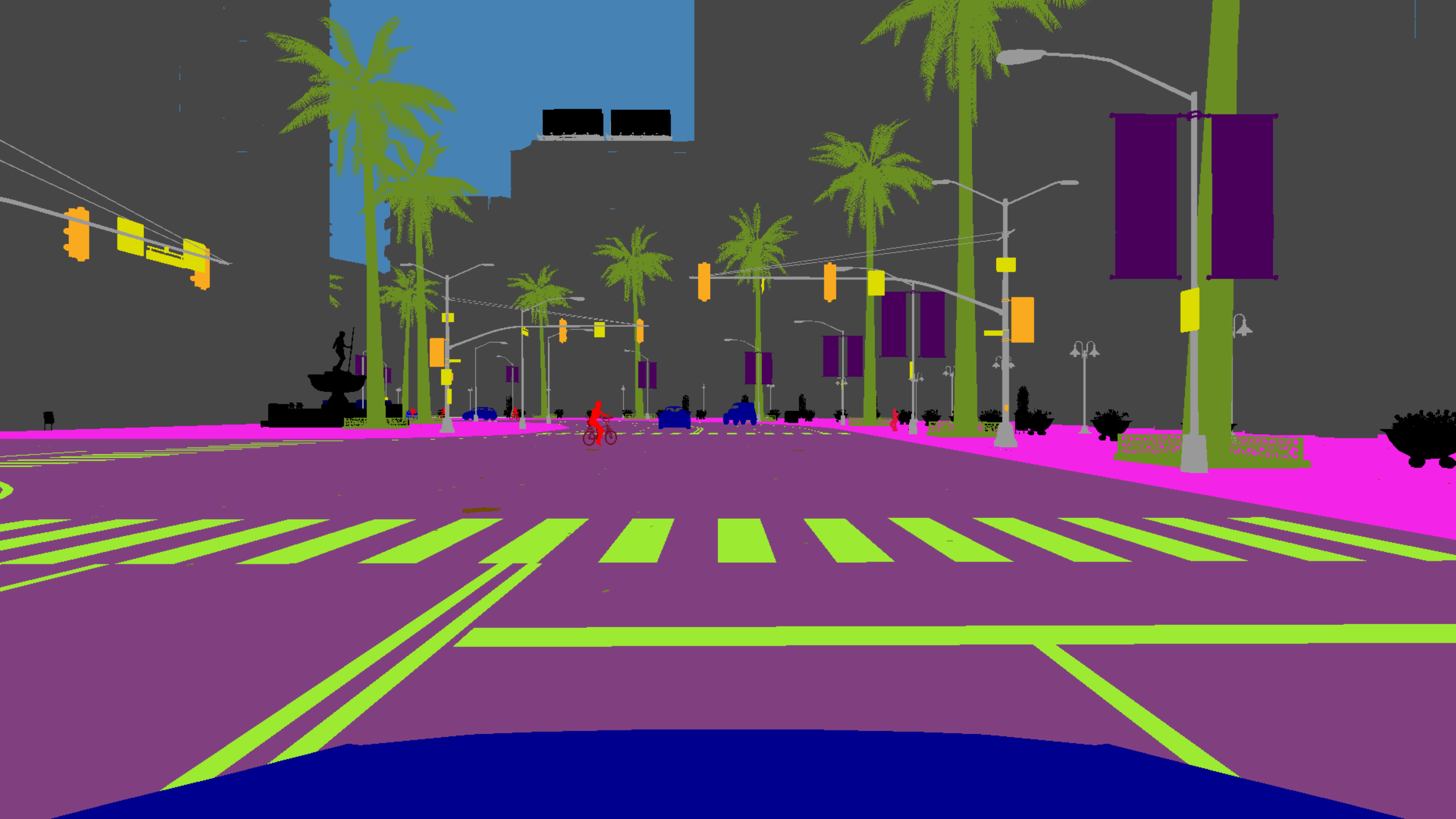}
        \end{subfigure}
        \subcaption{}
    \end{minipage}
    \begin{minipage}{.24\linewidth}
        \begin{subfigure}{\linewidth}
            \includegraphics[width=\linewidth]{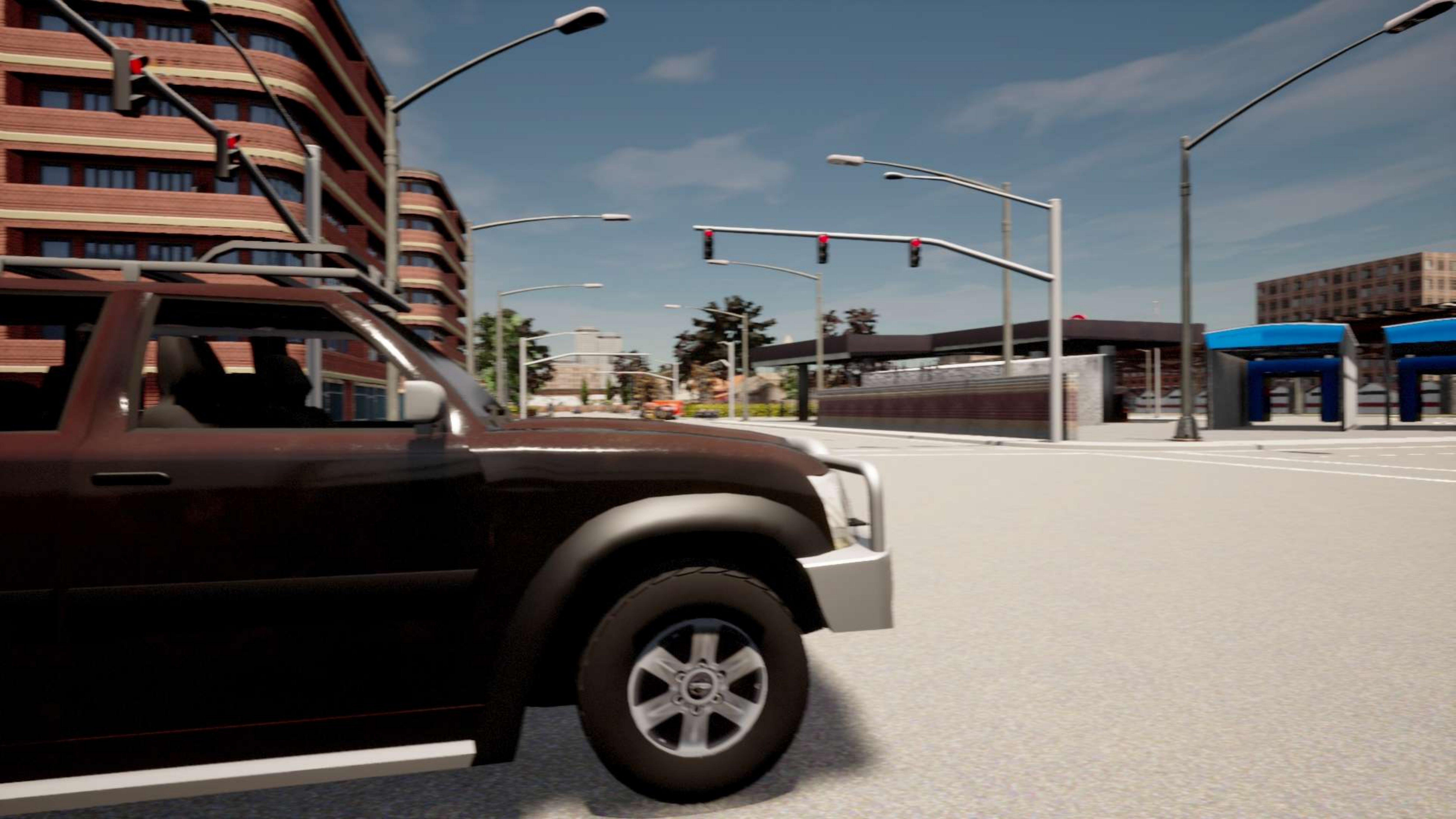}
        \end{subfigure}
        \begin{subfigure}{\linewidth}
            \includegraphics[width=\linewidth]{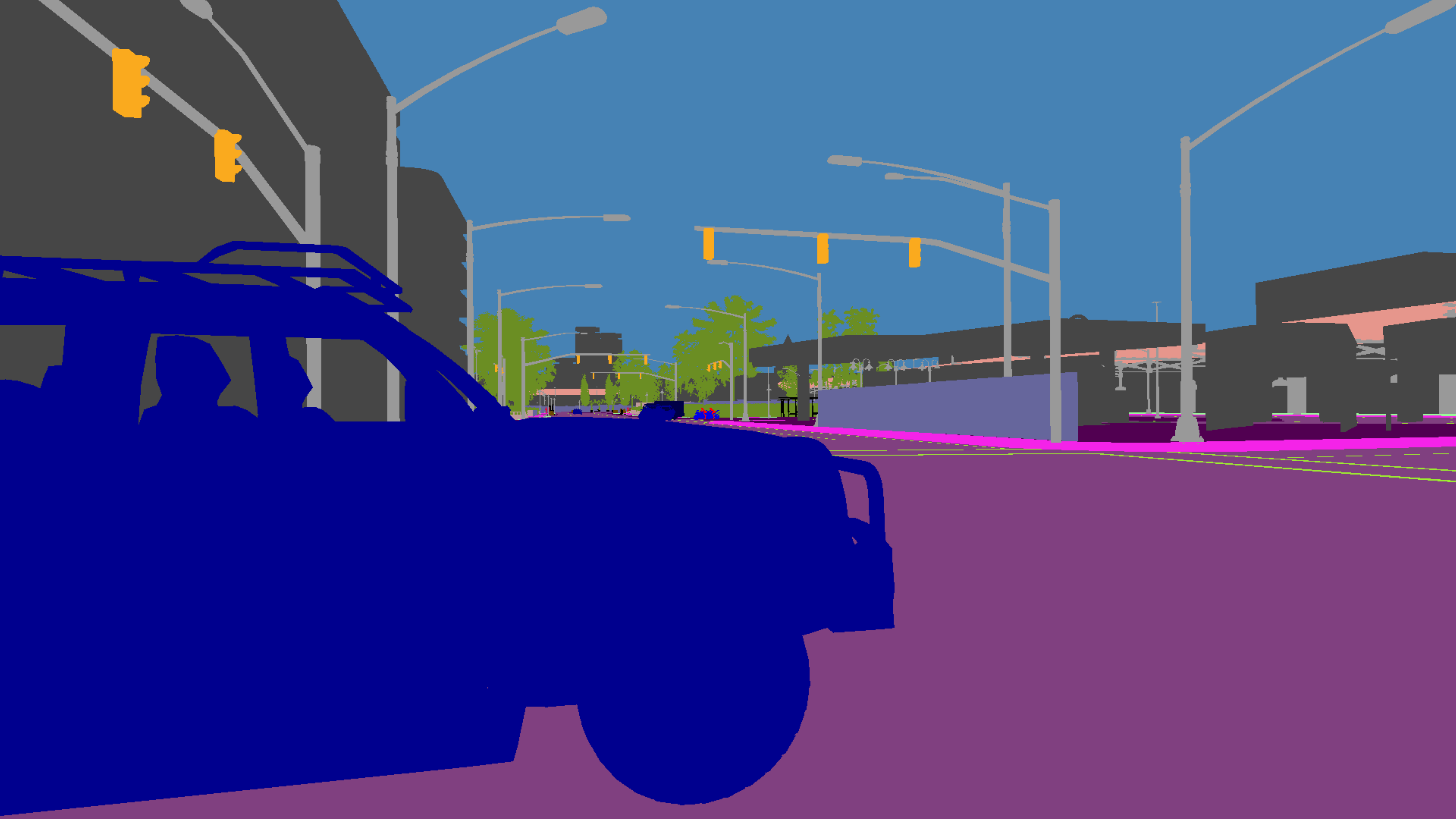}
        \end{subfigure}
        \subcaption{}
    \end{minipage}
    \begin{minipage}{.24\linewidth}
        \begin{subfigure}{\linewidth}
            \includegraphics[width=\linewidth]{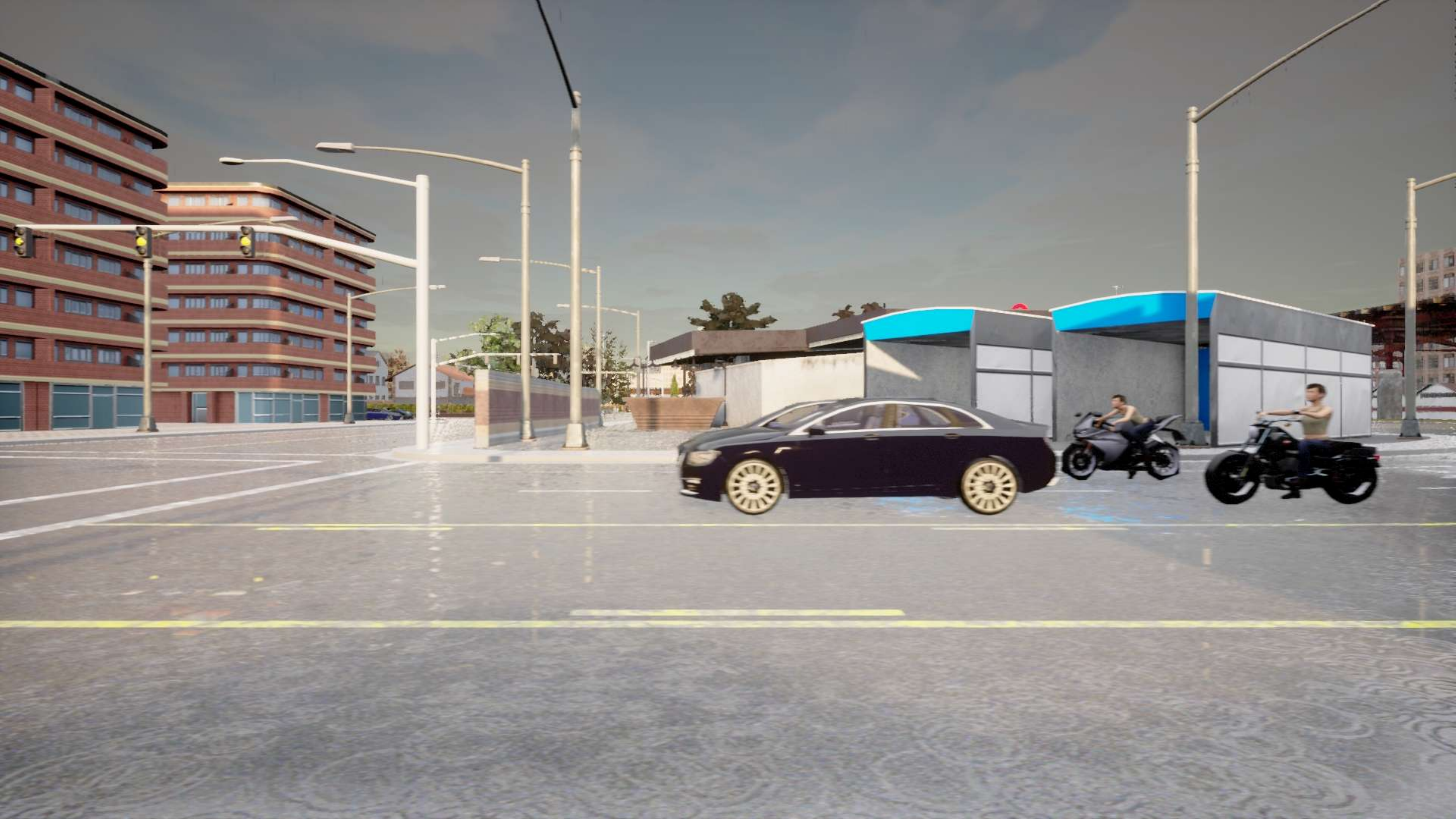}
        \end{subfigure}
        \begin{subfigure}{\linewidth}
            \includegraphics[width=\linewidth]{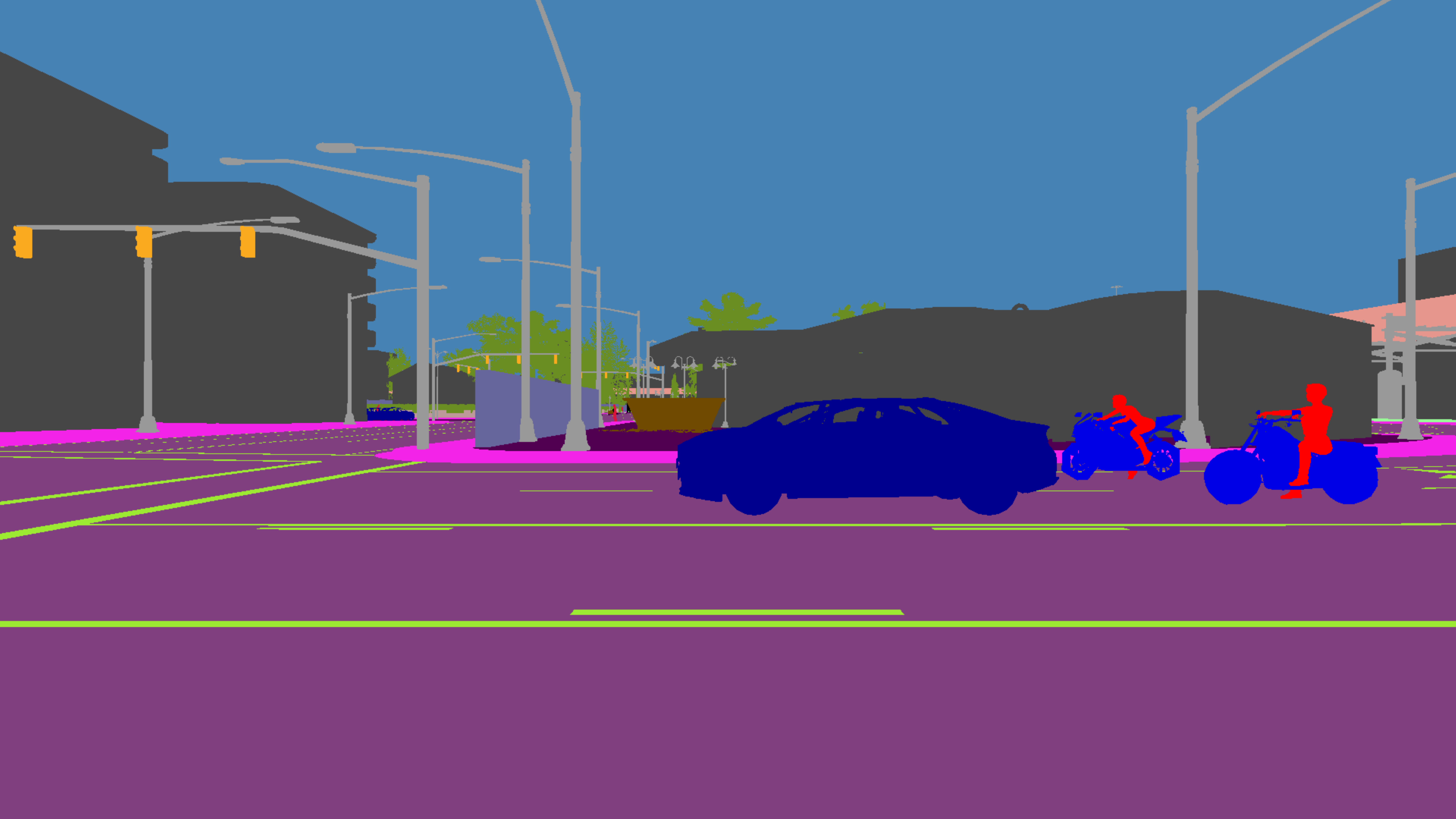}
        \end{subfigure}
        \subcaption{}
    \end{minipage}
    \vspace{2pt}
    \caption{a-b) show examples of the Town10 gallery and query sets from the front view, while c-d) are from the Town03 left view. Gallery and query sets are collected under the Clear Noon and Hard Rain Sunset weather conditions, respectively.
    }
    \label{fig:idda}
\end{figure*}

This new dataset was inspired by IDDA \cite{Alberti_RAL2020}, which was built from the CARLA virtual simulator \cite{Dosovitskiy_2017} specifically for semantic segmentation but without GPS annotations.
Following the methodology from \cite{Alberti_RAL2020} we used CARLA 0.9.10 to build a new dataset that includes both GPS/IMU information and pixel-wise semantic annotations with 25 semantic classes (with 17 of them in common to the Cityscapes~\cite{Cordts_CVPR2016} standard). 
This new synthetic dataset contains more than 40000 images (10091 per scenario) captured across two different urban maps (Town03 and Town10 from CARLA notation) and in two weather conditions, Clear Noon and Hard Rain Sunset. 
To collect the data we equipped the ego-vehicle with four cameras (front view, rear view, left view and right view).
We split the front and rear view frames captured in the Town10 in a gallery and a query set; the first including the Clear Noon images and the latter the Hard Rain Sunset ones, see \cref{fig:idda} a-b). 
The left and right view images captured in the Town03 are used as a validation set, following the same criterion used to split the training set, see \cref{fig:idda} c-d).

\vspace{-10pt}
\subsection{Synthetic-to-real domain adaptation}
\label{sec:unsupda_branch}
\noindent
The new synthetic dataset provides the training data for the method presented in \cref{sec:method}. However, there is a significant gap between the images from this simulator (\emph{source domain} $X_s$) and from the real-world (\emph{target domain} $X_t$).
To reduce this gap we use adversarial training that aims at aligning the features extracted from the synthetic and target domains. 
We assume having available at training time a set of unlabeled target domain images, besides the labeled synthetic data. 
While the VPR and SemSeg tasks illustrated in \cref{sec:vpr_branch,sec:semseg_branch} are trained using only the synthetic data $X_s$, we introduce a domain discriminator $D$ similar to \cite{Radford_ICLR2016} to distinguish the source features produced by $D_{seg}$ from the target ones. The discriminator is trained with the binary cross-entropy loss:
\begin{equation}
\begin{aligned}
    \mathcal{L}_{discr} = - \sum_{i \in \mathcal{I}} (1-z)logD(p_i(x_s)) + z log D(p_i(x_t)) 
\end{aligned}
\label{eq:loss_discr}
\end{equation}
with $x_s \in X_s$ and $x_t \in X_t$, $p_i(x)$ the features from $D_{seg}$ and z a parameter which is 0 if the features are from the source and 1 if the features are from the target domain. 
Overall, the adversarial training tries to fool the encoder and the SemSeg decoder, computing a binary cross-entropy loss on $\mathcal{X}_t$ labeled like source:
\begin{equation}
    \mathcal{L}_{adv} = 
    - \frac{1}{|\mathcal{I}|} \sum_{i \in \mathcal{I}} logD(p_i(x_t))
    \label{eq:loss_adv}
\end{equation}
In this way, the model learns to align the features for every domain and the encoder becomes able to extract discriminative local embeddings for the real world VPR task.

Summarizing, the overall training loss function becomes:
\begin{equation}
    \mathcal{L}_{tot} = \mathcal{L}_{VPR-SemSeg} + \beta \cdot \mathcal{L}_{adv} + \gamma \cdot \mathcal{L}_{discr}
    \label{eq:loss_tot}
\end{equation}    
where $\beta$ and $\gamma$ are scalar weights and $\mathcal{L}_{adv}$ affects the encoder weights.

\section{Experiments}
\vspace{-10pt}
\myparagraph{Comparisons with other methods.}
We assess the effectiveness of our approach comparing it to GeM \cite{Radenovic_PAMI2019} and RMAC \cite{Tolias_ICLR2016}, two state-of-the-art VPR methods that use global descriptors.
We also compare to DASGIL \cite{Hanjiang_TIP2021} which, similarly to our method, uses semantic information albeit selecting it in a top-down manner. 
Besides semantics, DASGIL can also leverage depth information to further enhance the descriptors. 
We compare to three different versions of DASGIL: the pre-trained models released by the authors using their proposed Flatten Discriminator (DASGIL\_FD $\bullet$) and Cascade Discriminator (DASGIL\_CD $\bullet$), as well as the model with Flatten Discriminator trained by us on our dataset but without depth information (DASGIL\_nodepth). DASGIL\_FD $\bullet$ and DASGIL\_CD $\bullet$ are pretrained on KITTI and VIRTUAL KITTI 2, while just DASGIL\_nodepth is pretrained on our novel dataset. 
In the comparison with DASGIL we must also note that it extracts and concatenates the local features from the i-th and j-th \emph{conv} layer of the shared encoder, producing final embeddings of variable dimension $1 \times 64 \times (H_i \times W_i + H_j \times W_j)$. On RobotCar \cite{Maddern_IJRR2017}), this results in $15360$-D (dimensional) descriptors \wrt the more compact $3072$-D descriptors produced by our method. 

Concerning the implementation of our method, the details are provided in the supplementary material.

\myparagraph{Datasets and protocol.}
For the evaluation we used Oxford RobotCar \cite{Maddern_IJRR2017} as the inference dataset. 
Oxford RobotCar is a collection of images from the city of Oxford taken in different environmental conditions from a car-mounted camera. We use the Overcast scenario as the gallery, while the queries are divided into four scenarios: Rain, Snow, Sun, and Night, with one image sampled every 5 meters in order to decrease data redundancy.
In order to have a fair comparison and truly assess the effectiveness of our solution, which cannot be trained directly on the target domain because of the lack of semantic labels, we trained all models on our new synthetic dataset, adding to GeM and RMAC the same unsupervised domain adaptation (DA) module used in our network. On other hand, DASGIL, which also uses semantic labels, already includes a DA module.
Despite the usage of DA techniques we can expect the result to be considerably lower than if the models would be trained directly on the target domain, due to the strong domain shift. For completeness, in the supplementary material we show some additional comparisons and also results of the models trained on the real-world target domain, using imperfect semantic labels generated by an expert network.
Finally, we also compare the generalization capability of these methods, trained on the synthetic dataset with Oxford RobotCar as target domain, and testing them on three other datasets: Pitts30k \cite{Torii_2015PAMI}, the revisited version of Tokyo24/7~\cite{Torii_CVPR2015} (RTokyo) proposed in the supplementary material of \cite{Warburg_2020_CVPR}, and the recent MapillarySLS ~\cite{Warburg_2020_CVPR} validation set (since the test set labels have not yet been released). 
In all experiments we used the standard VPR metric Recall@N \cite{Arandjelovic_CVPR2016}, considering a retrieved gallery image as positive if it is within 25 meters from the query.

\begin{table*}[ht]
\centering
\begin{adjustbox}{width=\textwidth}

\begin{tabular}{r|c|c|c|c|c|c} 
\hline
Method               & \multicolumn{1}{c}{\begin{tabular}[c]{@{}c@{}}Overcast/Rain\\ 1 / 5 / 10\end{tabular}} & \multicolumn{1}{c}{\begin{tabular}[c]{@{}c@{}}Overcast/Snow\\ 1 / 5 / 10\end{tabular}} & \multicolumn{1}{c}{\begin{tabular}[c]{@{}c@{}}Overcast/Sun\\ 1 / 5 / 10\end{tabular}} & \begin{tabular}[c]{@{}c@{}}Overcast/Night\\ 1 / 5 / 10\end{tabular} & \begin{tabular}[c]{@{}c@{}}Avg\\ 1 / 5 / 10\end{tabular} & Avg            \\ 
\hline
GeM                  & 71.4/ 83.8 / 87.9                                                                      & 37.6 / 55.1 / 63.6                                                                     & 30.9 / 47.9 / 55.7                                                                    & 4.4 / 12.2 / 18.3                                                   & 36.1 / 49.7 / 56.4                                       & 47.4           \\
GeM  + DA            & 73.6 / 87.4 / 91.2                                                                     & 41.2 / 59.9 / 68.6                                                                     & 32.6 / 52.0 / 60.4                                                                    & 7.3 / 20.5 / 29.4                                                   & 38.7 / 54.9 / 62.4                                       & 52.0           \\
GeM $\star$          & 69.6 / 83.8 / 88.3                                                                     & 40.3 / 58.4 / 66.3                                                                     & 32.7 / 51.2 / 61.1                                                                    & 6.0 / 16.5 / 23.6                                                   & 37.2 / 52.5 / 59.8                                       & 49.8           \\
GeM + DA $\star$     & 78.7 / 90.8 / 93.7                                                                     & 43.8 / 64.7 / 72.8                                                                     & 33.8 / 54.3 / 64.1                                                                    & 10.2 / 25.2 / 34.2                                                  & 41.6 / 58.7 / 66.2                                       & 55.5           \\
RMAC                 & 73.9 / 86.9 / 90.9                                                                     & 35.1 / 53.5 / 61.6                                                                     & 22.7 / 39.9 / 48.0                                                                    & 2.5 / 8.2 / 13.0                                                    & 33.6 / 47.1 / 53.4                                       & 44.7           \\
RMAC  + DA           & 69.8 / 83.2 / 88.0                                                                     & 39.8 / 58.4 / 66.5                                                                     & 24.9 / 42.6 / 51.7                                                                    & 3.9 / 10.3 / 16.3                                                   & 34.6 / 48.6 / 55.6                                       & 46.3           \\
RMAC $\star$         & 68.6 / 82.5 / 87.0                                                                     & 35.6 / 52.8 / 60.3                                                                     & 27.0 / 43.4 / 52.6                                                                    & 2.9 / 10.0 / 15.8                                                   & 33.5 / 47.2 / 53.9                                       & 44.9           \\
RMAC + DA $\star$    & 65.3 / 80.0 / 84.6                                                                     & 32.0 / 52.1 / 61.6                                                                     & 25.6 / 44.0 / 54.2                                                                    & 3.3 / 9.5 / 14.9                                                    & 31.5 / 46.4 / 53.8                                       & 43.9           \\
DASGIL\_nodepth      & 87.2 / 92.9 / 94.5                                                                     & 39.3 / 55.3 / 63.1                                                                     & 20.8 / 33.3 / 40.6                                                                    & 6.5 / 14.2 / 19.5                                                   & 38.4 / 48.9 / 54.4                                       & 47.3           \\
DASGIL\_FD $\bullet$ & 76.8 / 83.7 / 86.4                                                                     & 53.4 / 65.0 / 70.2                                                                     & 46.5 / 59.3 / 65.0                                                                    & 2.0 / 6.3 / 11.4                                                    & 44.7 / 53.6 / 58.3                                       & 52.2           \\
DASGIL\_CD $\bullet$ & 75.3 / 84.7 / 87.9                                                                     & 34.0 / 45.7 / 51.9                                                                     & 12.0 / 21.7 / 28.0                                                                    & 1.0 / 4.1 / 6.4                                                     & 30.6 / 39.1 / 43.6                                       & 37.7           \\ 
\hline
Ours w DL            & \textbf{92.2 / 96.8 / 97.8}                                                            & \textbf{68.0 / 83.4 / 89.0}                                                            & \textbf{62.6 / 79.6 / 85.0}                                                           & 11.3 / 24.5 / 34.1                                                  & \textbf{58.5} / 71.1 / 76.5                              & \textbf{68.7}  \\
Ours w DL $\star$    & 90.5 / 96.9 / 97.7                                                                     & 62.5 / 79.6 / 85.7                                                                     & 55.1 / 73.7 / 81.0                                                                    & 10.8 / 25.3 / 34.7                                                  & 54.7 / 68.9 / 74.8                                       & 66.1           \\
Ours w PSP           & 90.7 / 96.5 / 97.3                                                                     & 59.5 / 75.5 / 82.4                                                                     & 47.0 / 65.2 / 71.9                                                                    & 12.0 / 26.4 / 34.7                                                  & 52.3 / 65.9 / 71.6                                       & 63.3           \\
Ours w PSP $\star$   & 91.2 / 96.4 / 97.7                                                                     & 61.4 / 83.1 / 88.4                                                                     & 55.4 / 74.0 / 81.0                                                                    & \textbf{14.2 / 32.6 / 42.3}                                         & 55.6 / \textbf{71.5 / 77.4}                              & 68.2           \\
\hline
\end{tabular}

\end{adjustbox}
\vspace{2pt}
\caption{Results on the Oxford RobotCar \cite{Maddern_IJRR2017}, using Overcast as gallery. $\star$ means ResNet101 as encoder, all the others are with ResNet50. DA stays for Domain Adaptation, DL is for DeepLabv2 \cite{Chen2018DeepLabSI} segmentation decoder, while PSP is for PSPNet \cite{Zhao_CVPR2017}. $\bullet$ indicates the experiments of \cite{Hanjiang_TIP2021} with the trained models provided by the authors.}
\label{tab:results_robotcar}
\vspace{-2pt}
\end{table*}

\vspace{-35pt}
\subsection{Results}
\vspace{-1pt}
\noindent
The results of the experiments are reported in \cref{tab:results_robotcar}.
We observe that all methods perform worse when tested on queries from the Night and Sun scenarios, due to a stronger visual dissimilarity with respect to the gallery images that are taken from the Overcast scenario. 

Overall, we see that our architecture largely outperforms all other methods, with and without DA, from a minimum of 8\% to a maximum of 24\%. Curiously, with RMAC \cite{Tolias_ICLR2016} results seem to get worse when adding DA. We do not have a definitive explanation for this negative effect, but we confirmed it with the generalization experiments presented in \cref{tab:results_gen}.
Even DASGIL, which is the most similar method to our work, \cref{tab:results_robotcar} struggles to generalize to all scenarios. The model trained on our datasets and without the depth information (DASGIL\_nodepth) performs very well on the Rain scenario, but it still results on average worse than the GeM. We suspect that this method suffers considerably from the large domain gap between our synthetic dataset and the RobotCar \cite{Maddern_IJRR2017} target images. For this reason we extend the experiments by using also the pretrained models provided by the authors (DASGIL\_FD $\bullet$ and DASGIL\_CD $\bullet$), that are trained on synthetic and real datasets Virtual KITTI 2 \cite{cabon2020virtual} and KITTI \cite{Geiger_CVPR2012}. Nevertheless, these implementations still remain lower than our method by at least 11\%. We further confirm our assumptions in \cref{tab:results_gen} testing the generalization capability on very large city datasets.
Finally, the generalization results shown in \cref{tab:results_gen} demonstrate that our solution generalizes to unseen domains better than all the other methods by usually a large margin. Qualitative results are provided in the supplementary material.

\begin{table*}[ht]
\centering
\begin{adjustbox}{width=\textwidth}
\begin{tabular}{r|c|c|c|c|c} 
\hline
Method               & \multicolumn{1}{c}{\begin{tabular}[c]{@{}c@{}}Pitts30k - Val\\ 1 / 5 / 10\end{tabular}} & \multicolumn{1}{c}{\begin{tabular}[c]{@{}c@{}}Pitts30k - Test\\ 1 / 5 / 10\end{tabular}} & \multicolumn{1}{c}{\begin{tabular}[c]{@{}c@{}}RTokyo - Val\\ 1 / 5 / 10\end{tabular}} & \multicolumn{1}{c}{\begin{tabular}[c]{@{}c@{}}RTokyo - Test\\ 1 / 5 / 10\end{tabular}} & \begin{tabular}[c]{@{}c@{}}MapillarySLS \\ 1 / 5 / 10\end{tabular}  \\ 
\hline
GeM                  & 37.6 / 56.5 / 64.9                                                                      & 39.5 / 59.1 / 67.3                                                                       & 28.7 / 43.5 / 51.2                                                                    & 8.1 / 17.4 / 24.8                                                                      & 26.5 / 38.6 / 44.5                                                  \\
GeM  + DA            & 41.1 / 61.0 / 69.0                                                                      & 41.7 / 61.6 / 69.7                                                                       & 33.7 / 49.4 / 57.0                                                                    & 13.4 / 25.7 / 31.6                                                                     & 27.2 / 40.0 / 46.1                                                  \\
GeM $\star$          & 38.0 / 57.9 / 67.4                                                                      & 38.8 / 58.5 / 67.3                                                                       & 26.4 / 41.5 / 49.2                                                                    & 10.1 / 20.4 / 27.4                                                                     & 25.7 / 39.5 / 45.0                                                  \\
GeM + DA $\star$     & 47.1 / 68.9 / 77.3                                                                      & 47.7 / 68.4 / 75.9                                                                       & 36.0 / 52.6 / 60.3                                                                    & 17.6 / 34.6 / 43.5                                                                     & 28.9 / 41.5 / 48.4                                                  \\
RMAC                 & 39.4 / 58.5 / 67.3                                                                      & 43.3 / 62.9 / 71.1                                                                       & 40.1 / 55.7 / 62.6                                                                    & 13.8 / 28.3 / 36.9                                                                     & 33.6 / 44.3 / 50.6                                                  \\
RMAC  + DA           & 37.8 / 57.2 / 66.1                                                                      & 41.9 / 62.1 / 71.3                                                                       & 36.3 / 51.7 / 59.0                                                                    & 10.4 / 23.8 / 31.5                                                                     & 28.8 / 40.4 / 46.7                                                  \\
RMAC $\star$         & 33.8 / 53.5 / 63.0                                                                      & 37.7 / 57.4 / 66.9                                                                       & 31.7 / 46.9 / 54.6                                                                    & 10.0 / 20.9 / 27.7                                                                     & 31.5 / 44.0 / 48.5                                                  \\
RMAC + DA $\star$    & 32.1 / 51.3 / 60.6                                                                      & 36.4 / 56.3 / 65.3                                                                       & 30.5 / 45.8 / 53.4                                                                    & 14.8 / 25.7 / 33.0                                                                     & 30.2 / 41.4 / 46.2                                                  \\
DASGIL\_nodepth      & 11.1 / 17.6 / 21.6                                                                      & 12.8 / 20.6 / 26.0                                                                       & 17.7 / 28.7 / 35.0                                                                    & 2.0 / 3.8 / 6.4                                                                        & 11.1 / 16.6 / 19.7                                                  \\
DASGIL\_FD $\bullet$ & 8.3 / 12.6 / 15.3                                                                       & 8.7 / 13.8 / 16.6                                                                        & 5.9 / 9.1 / 11.9                                                                      & 0.0 / 0.7 / 1.0                                                                        & 6.4 / 8.6 / 10.1                                                    \\
DASGIL\_CD $\bullet$ & 6.8 / 10.7 / 13.9                                                                       & 8.5 / 12.3 / 15.8                                                                        & 7.2 / 11.9 / 15.3                                                                     & 0.3 / 1.0 / 1.0                                                                        & 7.7 / 11.4 / 14.1                                                   \\ 
\hline
Ours w DL            & 56.3 / 73.8 / 80.1                                                                      & 58.9 / 75.2 / 80.4                                                                       & 49.6 / 64.1 / 70.3                                                                    & 20.7 / 37.1 / 45.5                                                                     & \textbf{34.6} / 45.8 / 52.3                                         \\
Ours w DL $\star$    & \textbf{57.9} / 76.3 / 82.8                                                             & \textbf{59.4} / 76.2 / 81.5                                                              & 49.0 / 64.0 / 70.3                                                                    & 21.4 / 38.0 / 45.0                                                                     & 34.5 / 46.8 / \textbf{53.0}                                         \\
Ours w PSP           & 52.9 / 70.9 / 77.9                                                                      & 56.3 / 73.0 / 79.0                                                                       & 47.3 / 62.1 / 68.6                                                                    & 17.4 / 30.9 / 38.6                                                                     & 33.0 / \textbf{47.2} / \textbf{53.0}                                \\
Ours w PSP $\star$   & 57.7 / \textbf{76.4} / \textbf{83.3}                                                    & 59.1 / \textbf{76.6} / \textbf{82.2}                                                     & \textbf{51.3} / \textbf{66.2} / \textbf{72.5}                                         & \textbf{26.9} / \textbf{44.6} / \textbf{51.9}                                          & 32.9 / 46.1 / 51.9                                                  \\
\hline
\end{tabular}
\end{adjustbox}
\vspace{2pt}
\caption{Results using Pitts30k \cite{Torii_2015PAMI} and RTokyo \cite{Warburg_2020_CVPR} validation and test sets, and MapillarySLS  \cite{Warburg_2020_CVPR} validation cities. * means ResNet101 as encoder while all the others are with ResNet50. DA stays for Domain Adaptation, DL means DeepLabv2 \cite{Chen2018DeepLabSI} segmentation decoder and PSP is for PSPNet \cite{Zhao_CVPR2017}. $\bullet$ indicates the experiments of \cite{Hanjiang_TIP2021} with the trained models provided by the authors.}
\label{tab:results_gen}
\vspace{-2pt}
\end{table*}


\vspace{-35pt}
\subsection{Ablation studies}
\vspace{-2pt}
We perform extensive experiments to assess the impact of the various modules in our architecture. Due to lack of space, these experiments and the related discussion of results are reported in the supplementary material.
\vspace{-5pt}
\section{Conclusions}
\vspace{-5pt}
\noindent
We have presented a new method for generating global descriptors for VPR, exploiting both visual appearance and semantic features at different scales. Our solution is founded on the intuition that not all the semantic content is useful for VPR. Unlike previous works that select the semantic information in a top-down manner, we let the model determine what semantic information to use, in a data driven way. The key for this, is an attention mechanism that lets the VPR guide the semantic segmentation. Experiments on well-known VPR benchmarks, where we surpass the current state-of-the-art methods, validate our intuition and architecture. We also show that our model generalizes well to unseen target domains. Finally, we contribute a new dataset, rich of RGB images under different conditions, pixel-wise semantic masks and GPS coordinate, that is instrumental to explore the connection between semantics and appearance in VPR task, and that we believe will be useful to the research community. Following acceptance, we will make the dataset and code publicly available.

\newpage
\section{Supplementary Material}
\subsection{Implementation details.}
The experiments with our architecture reported in the main paper are conducted  using the ResNet50 and the ResNet101 (pretrained on ImageNet) as the shared encoder, which is truncated before the last average pooling. We trained our method with DeepLabv2 \cite{Chen2018DeepLabSI} and PSPNet \cite{Zhao_CVPR2017} as SemSeg decoder. Training is performed using the 17 classes in common with the Cityscapes standard in semantic segmentation. The domain discriminator consists of 5 convolutional layers with kernel $4 \times 4$, stride 2, padding 1 and channel numbers \{64, 128, 256, 512, 1\}. Each layer, except the last one is followed by a Leaky ReLU activation function with a negative slope of 0.2. The SemSeg decoder is initialized with the normal distribution, while the multi-scale attention module with the Xavier initialization. The encoder, the SemSeg decoder and the multi-scale attention layer are trained with SGD with an initial learning rate of 1e-4. The domain discriminator is trained with Adam with an initial learning rate of 4e-4. The ”poly” learning rate decay with a power of 0.9, momentum 0.9 and weight decay to 0.0005 is used for all the modules. $\alpha$ and $\gamma$ are fixed to 0.5 while $\beta$ is set to 0.0005. We used the caching mechanisms as in \cite{Arandjelovic_CVPR2016} to refresh the gallery embeddings during the training. We pre-process the triplets with random-crops and horizontal-flips. The size of training images is 768x432. For fairness of comparison, all the experiments (baselines and ours) are validated on the source domain using the left/right views of Town3, not considering the accuracy on the real world images as a metric to stop the training. The overall network is trained end-to-end, while no image pre-processing and no whitening-PCA is performed for the evaluation phase.

\subsection{Ablation studies}
\myparagraph{Impact of different modules.}
We performed an extensive ablation study on the RobotCar \cite{Maddern_IJRR2017} scenarios
to evaluate the impact of each component of our method. For this study we used ResNet50 as encoder and DeepLabv2  \cite{Chen2018DeepLabSI} as SemSeg decoder.
As baseline, we consider the shared encoder followed by the GeM \cite{Radenovic_PAMI2019} pooling layer. Then, we added each component, \ie, the multi-scale attention block (Att), the segmentation decoder (SemSeg), the attention to guide the segmentation (G-SemSeg) and the unsupervised domain adaptation mechanism (DA), trying all combinations. 
As shown in \cref{{tab:ablation_extra}} each module improves the results, in particular across the single-scale embedding experiments even without the domain adaptation (DA) task, the semantic segmentation (SemSeg) and the guided configuration (G-SemSeg) provide strong boosts respectively of +10\% and +14\% on the average. 
Finally, we build the final architecture that adds up to an overall improvement of approximately 17\% over the baseline. 
Moreover, adding our novel multi-scale embeddings (ms-GeM) gives a further gain of 4\%.

\begin{table*}[t]
\centering
\begin{adjustbox}{width=\textwidth}

\begin{tabular}{ccccc|c|c|c|c|c|c} 
\hline
ms-GeM       & Att          & SemSeg       & G-SemSeg     & DA           & \multicolumn{1}{c}{\begin{tabular}[c]{@{}c@{}}Overcast/Rain\\ 1 / 5 / 10\end{tabular}} & \multicolumn{1}{c}{\begin{tabular}[c]{@{}c@{}}Overcast/Snow\\ 1 / 5 / 10\end{tabular}} & \multicolumn{1}{c}{\begin{tabular}[c]{@{}c@{}}Overcast/Sun\\ 1 / 5 / 10\end{tabular}} & \begin{tabular}[c]{@{}c@{}}Overcast/Night\\ 1 / 5 / 10\end{tabular} & \begin{tabular}[c]{@{}c@{}}Avg\\ 1 / 5 / 10\end{tabular} & Avg            \\ 
\hline
             &              &              &              &              & 71.4 / 83.8 / 87.9                                                                     & 37.6 / 55.1 / 63.6                                                                     & 30.9 / 47.9 / 55.7                                                                    & 4.4 / 12.2 / 18.3                                                   & 36.1 / 49.7 / 56.4                                       & 47.4           \\
             & $\checkmark$ &              &              &              & 71.4 / 83.7 / 88.0                                                                     & 40.6 / 59.3 / 67.2                                                                     & 31.1 / 50.3 / 59.1                                                                    & 4.9 / 15.1 / 21.9                                                   & 37.0 / 52.1 / 59.1                                       & 49.4           \\
             & $\checkmark$ & $\checkmark$ &              &              & 82.7 / 93.3 / 95.5                                                                     & 50.0 / 70.1 / 79.2                                                                     & 45.5 / 66.2 / 74.4                                                                    & 7.5 / 17.9 / 25.9                                                   & 46.4 / 61.9 / 68.8                                       & 59.0           \\
             & $\checkmark$ &              &              & $\checkmark$ & 86.8 / 93.8 / 96.2                                                                     & 56.1 / 73.6 / 80.6                                                                     & 43.8 / 64.0 / 72.6                                                                    & 6.5 / 16.9 / 25.4                                                   & 48.3 / 62.1 / 68.7                                       & 59.7           \\
             & $\checkmark$ & $\checkmark$ &              & $\checkmark$ & 81.4 / 92.3 / 94.8                                                                     & 53.6 / 73.3 / 81.3                                                                     & 44.1 / 64.6 / 72.9                                                                    & 10.2 / 24.6 / 34.7                                                  & 47.3 / 63.7 / 70.9                                       & 60.6           \\
             & $\checkmark$ & $\checkmark$ & $\checkmark$ &              & 87.6 / 95.1 / 96.7                                                                     & 56.3 / 75.7 / 83.0                                                                     & 50.3 / 69.6 / 84.1                                                                    & 8.8 / 20.2 / 29.2                                                   & 50.8 / 65.1 / 73.3                                       & 63.1           \\
             & $\checkmark$ & $\checkmark$ & $\checkmark$ & $\checkmark$ & 85.2 / 94.1 / 96.3                                                                     & 59.5 / 78.7 / 85.3                                                                     & 50.5 / 69.1 / 76.9                                                                    & \textbf{11.7 / 27.2 / 37.3}                                         & 51.7 / 67.3 / 73.9                                       & 64.3           \\ 
\hline
$\checkmark$ &              &              &              &              & 85.8 / 93.4 / 95.6                                                                     & 56.5 / 72.6 / 79.1                                                                     & 51.0 / 69.5 / 76.6                                                                    & 7.4 / 20.0 / 29.2                                                   & 50.2 / 63.9 / 70.1                                       & 61.4           \\
$\checkmark$ & $\checkmark$ &              &              &              & 90.5 / 96.3 / 97.5                                                                     & 59.2 / 75.9 / 82.6                                                                     & 50.1 / 67.8 / 75.2                                                                    & 5.8 / 16.4 / 25.7                                                   & 51.4 / 64.1 / 70.3                                       & 61.9           \\
$\checkmark$ & $\checkmark$ & $\checkmark$ &              &              & 91.1 / 96.0 / 97.0                                                                     & 55.6 / 72.7 / 79.5                                                                     & 54.8 / 72.2 / 79.2                                                                    & 7.1 / 17.4 / 25.5                                                   & 52.2 / 64.6 / 70.3                                       & 62.4           \\
$\checkmark$ & $\checkmark$ &              &              & $\checkmark$ & 87.6 / 94.8 / 96.6                                                                     & 60.3 / 78.0 / 84.1                                                                     & 48.9 / 68.7 / 75.7                                                                    & 7.9 / 20.2 / 29.0                                                   & 51.2 / 65.4 / 71.3                                       & 62.6           \\
$\checkmark$ & $\checkmark$ & $\checkmark$ &              & $\checkmark$ & 89.9 / 96.0 / 97.3                                                                     & 63.1 / 80.4 / 86.4                                                                     & 57.9 / 75.6 / 81.6                                                                    & 7.7 / 18.6 / 27.7                                                   & 54.6 / 67.6 / 73.3                                       & 65.2           \\
$\checkmark$ & $\checkmark$ & $\checkmark$ & $\checkmark$ &              & 91.5 / 96.4 / 97.3                                                                     & 58.4 / 74.6 / 80.6                                                                     & 53.9 / 72.2 / 78.9                                                                    & 7.9 / 20.8 / 29.8                                                   & 53.0 / 66.0 / 71.7                                       & 63.5           \\
$\checkmark$ & $\checkmark$ & $\checkmark$ & $\checkmark$ & $\checkmark$ & \textbf{92.2 / 96.8 / 97.8}                                                            & \textbf{68.0 / 83.4 / 89.0}                                                            & \textbf{62.6 / 79.6 / 85.0}                                                           & 11.3 / 24.5 / 34.1                                                  & \textbf{58.5 / 71.1 / 76.5}                              & \textbf{68.7}  \\ 
\hline\hline
             &              &              &              & $\checkmark$ & 73.6 / 87.4 / 91.2                                                                     & 41.2 / 59.9 / 68.6                                                                     & 32.6 / 52.0 / 60.4                                                                    & 7.3 / 20.5 / 29.4                                                   & 38.7 / 54.9 / 62.4                                       & 52.0           \\
             &              & $\checkmark$ &              &              & 83.3 / 92.7 / 95.0                                                                     & 46.3 / 66.1 / 75.4                                                                     & 40.7/ 60.9 / 69.0                                                                     & 8.1 / 18.8 / 26.7                                                   & 44.6 / 59.6 / 66.5                                       & 56.9           \\
             &              & $\checkmark$ &              & $\checkmark$ & 81.6 / 92.2 / 94.7                                                                     & 52.7 / 72.6 / 80.3                                                                     & 45.5 / 65.8 / 74.2                                                                    & 8.4 / 20.7 / 30.7                                                   & 47.1 / 62.8 / 70.0                                       & 60.0           \\ 
\hline
$\checkmark$ &              &              &              & $\checkmark$ & 85.7 / 94.1 / 95.9                                                                     & 59.6 / 76.9 / 83.6                                                                     & 45.8 / 67.3 / 74.8                                                                    & 5.2 / 16.4/ 25.4                                                    & 49.1 / 63.7 / 69.9                                       & 60.9           \\
$\checkmark$ &              & $\checkmark$ &              &              & 88.9 / 95.5 / 96.8                                                                     & 58.4 / 75.3 / 82.1                                                                     & 54.0 / 72.8 / 80.1                                                                    & 7.7 / 18.2 / 25.8                                                   & 52.3 / 65.5 / 71.2                                       & 63.0           \\
$\checkmark$ &              & $\checkmark$ &              & $\checkmark$ & 91.4 / 96.8 / 97.6                                                                     & 61.7 / 78.9 / 85.7                                                                     & 57.4 / 75.2 / 81.5                                                                    & 7.7 / 20.2 / 29.4                                                   & 54.6 / 67.8 / 73.6                                       & 65.3           \\
\hline
\end{tabular}

\end{adjustbox}
\vspace{2pt}
\caption{Ablation study to assess the contribution of each component in our architecture. \emph{ms-GeM} stays for Multi-Scale Embeddings obtained through our new multi-scale pooling layer, \emph{Att} indicates that the attention module is used in the VPR task, \emph{SemSeg} and \emph{G-SemSeg} denote that Semantic Segmentation and Guided Semantic Segmentation are active, while \emph{DA} stands for the Domain Adaptation task.}
\label{tab:ablation_extra}
\end{table*}

\begin{figure}[t!]
    \centering
    \includegraphics[width=\linewidth]{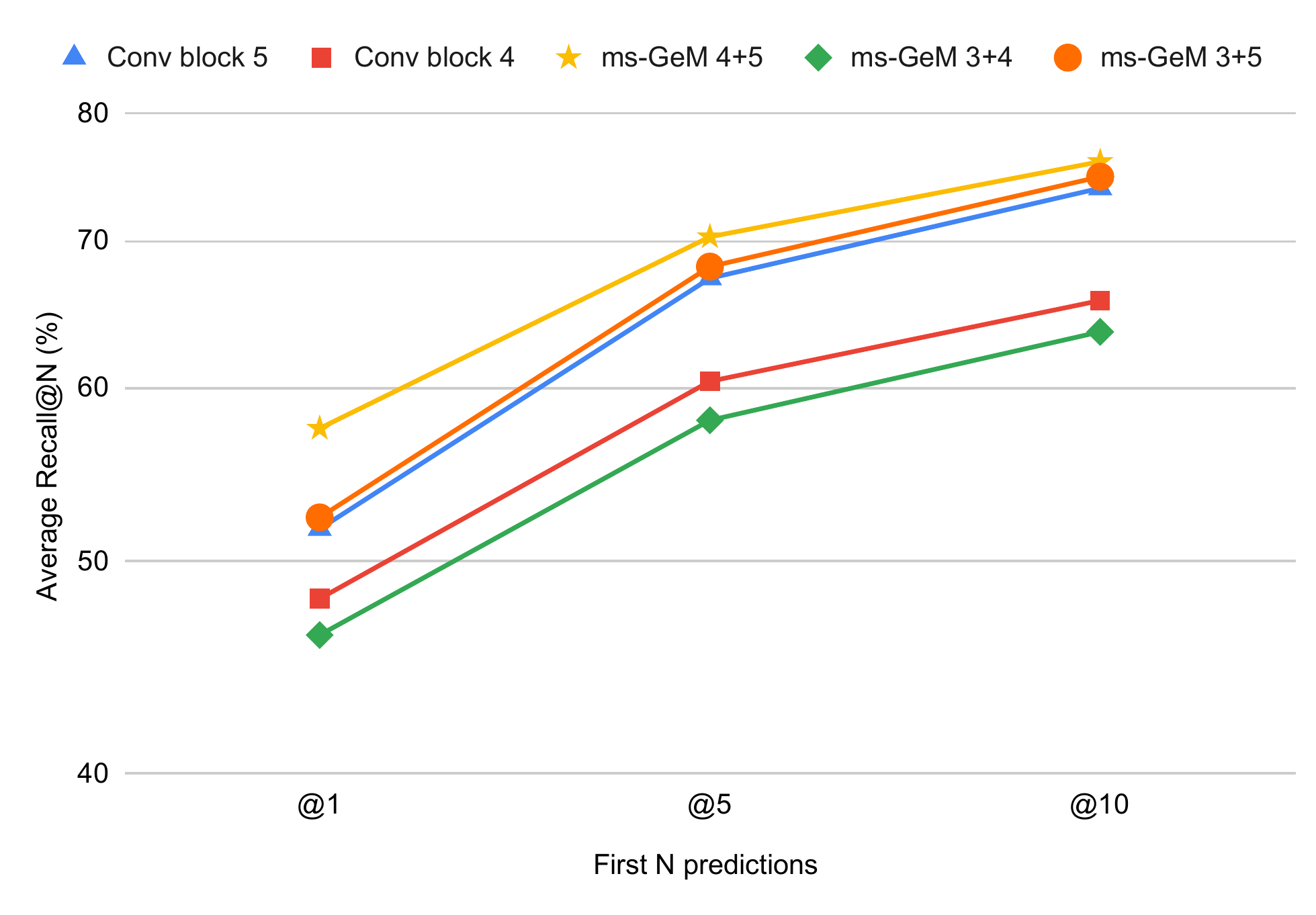}
    \caption{Average recalls across Rain, Snow, Sun and Night queries of RobotCar \cite{Maddern_IJRR2017}, getting local features at different scale.}
    \label{fig:ablation_multiscale}
\end{figure}

\myparagraph{Multi-scale embeddings.}
In \cref{fig:ablation_multiscale} we present an ablation analysis of the encoder local features that should be aggregated to produce robust embeddings. We get the best results when we use the local features from the last two \emph{conv} blocks (ms-GeM 4+5), which have a higher level of abstraction and are useful for distinguishing different classes. Local features from previous layers, on the other hand, create global embeddings that are inadequate to accurately represent the places.

\subsection{Qualitative results}
\begin{figure*}[t!]
    \centering
    \includegraphics[width=1.0\linewidth]{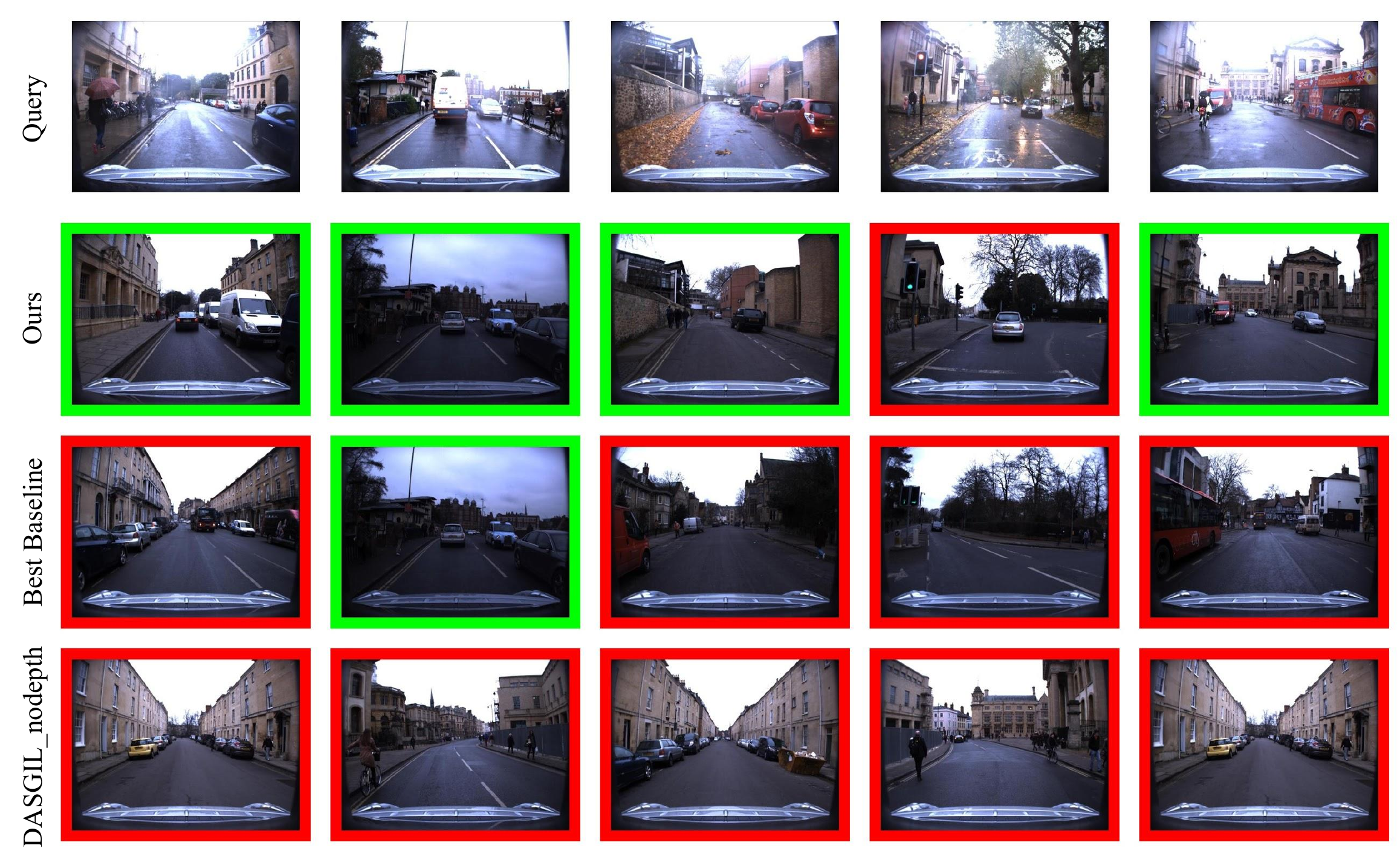}
    \caption{Qualitative predictions testing on Oxford RobotCar \cite{Maddern_IJRR2017} with queries from Rain scenario and gallery from Overcast scenario.}
    \label{fig:preds_rain}
\end{figure*}

\begin{figure*}[t!]
    \centering
    \begin{subfigure}{1.0\textwidth}
        \includegraphics[width=\linewidth]{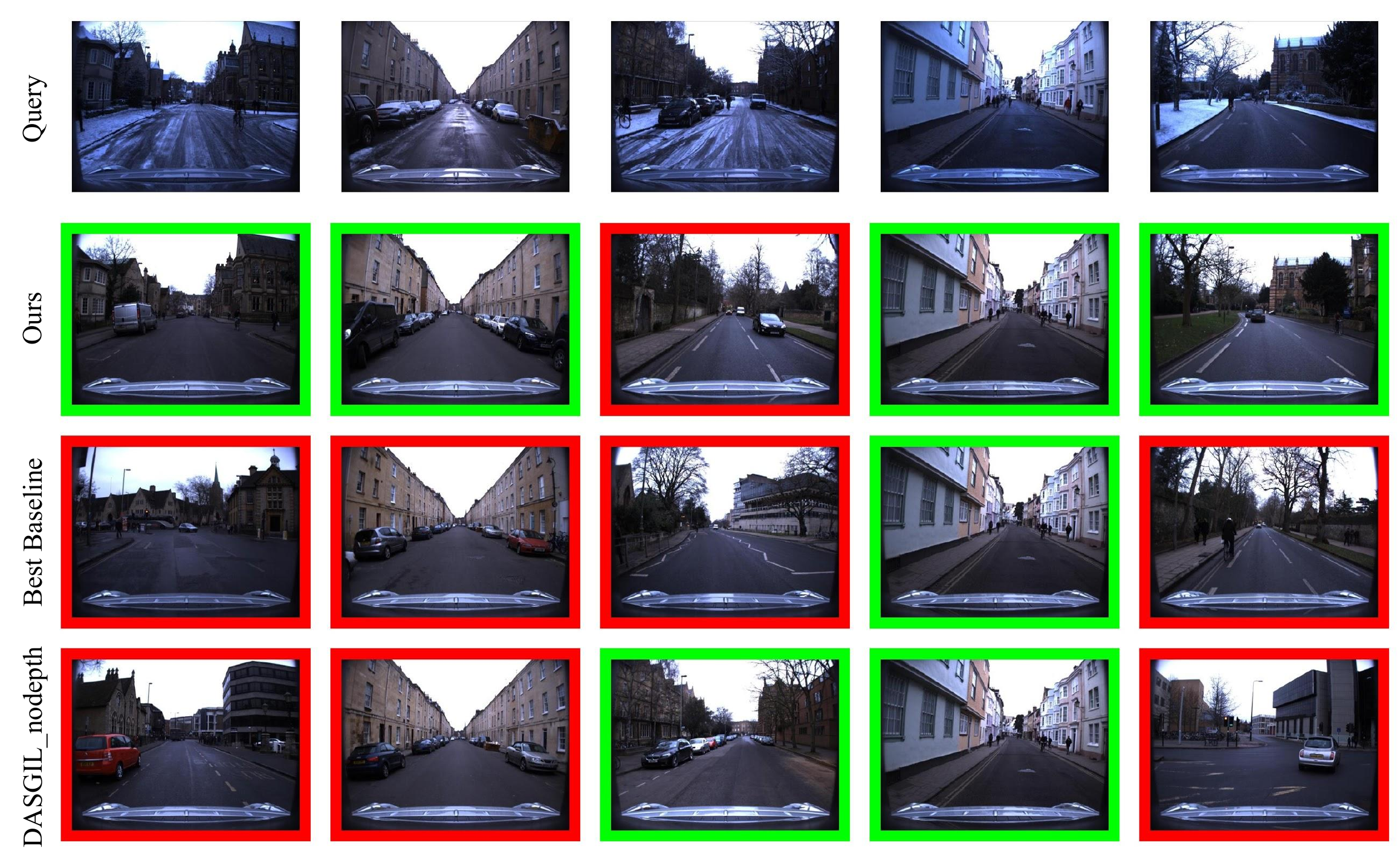}
        \caption{}
    \end{subfigure}
    \vspace{6pt}
    \hrule
    \vspace{6pt}
    \begin{subfigure}{1.0\textwidth}
        \includegraphics[width=\linewidth]{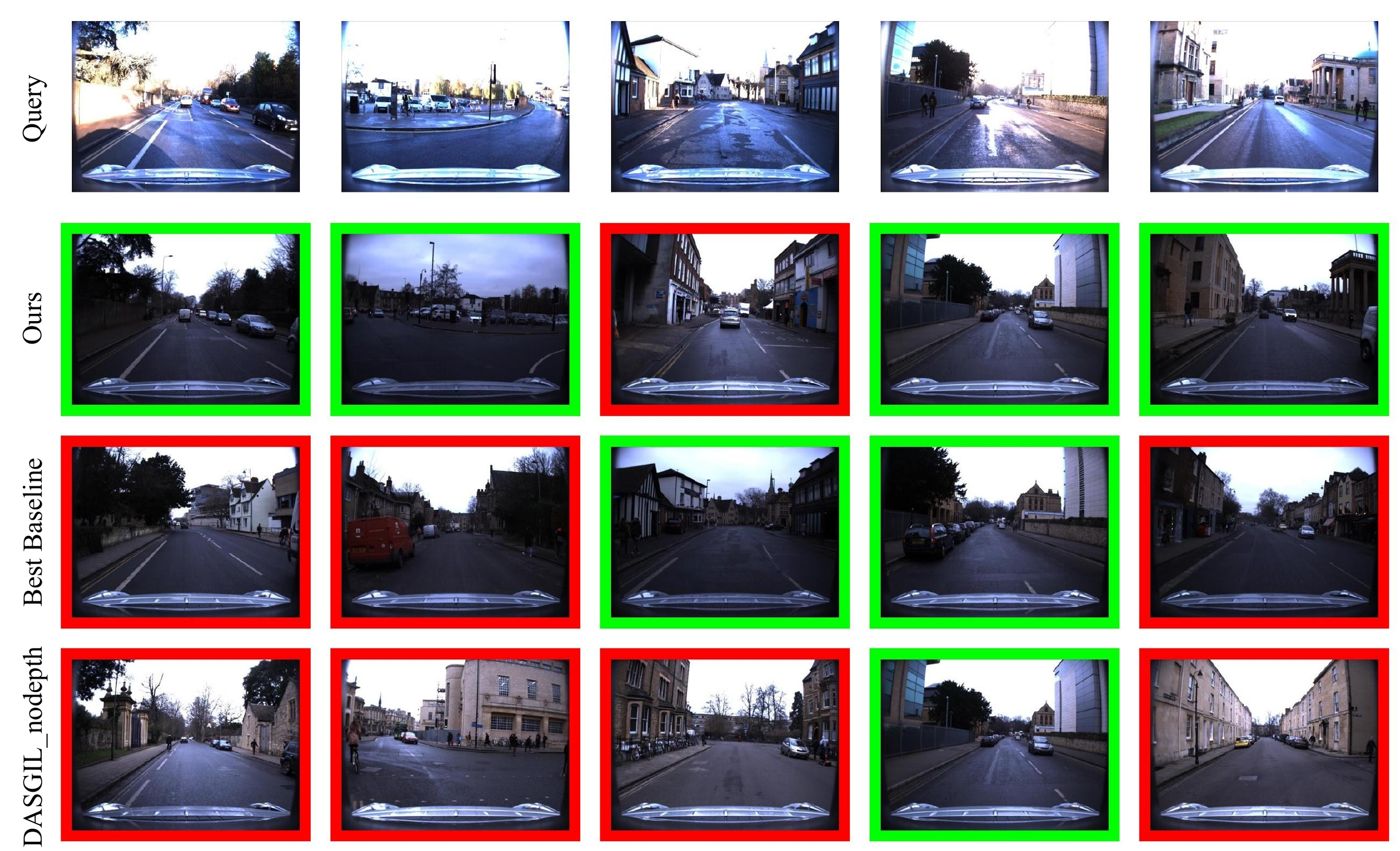}
        \caption{}
    \end{subfigure}
    \caption{Qualitative predictions testing on Oxford RobotCar \cite{Maddern_IJRR2017} with queries from Snow (a) and Sun (b) scenarios, while the gallery is always composed by the Overcast scenario.}
    \label{fig:preds_snow_and_sun}
\end{figure*}

\begin{figure*}[t!]
    \centering
    \begin{subfigure}{1.0\textwidth}
        \includegraphics[width=\linewidth]{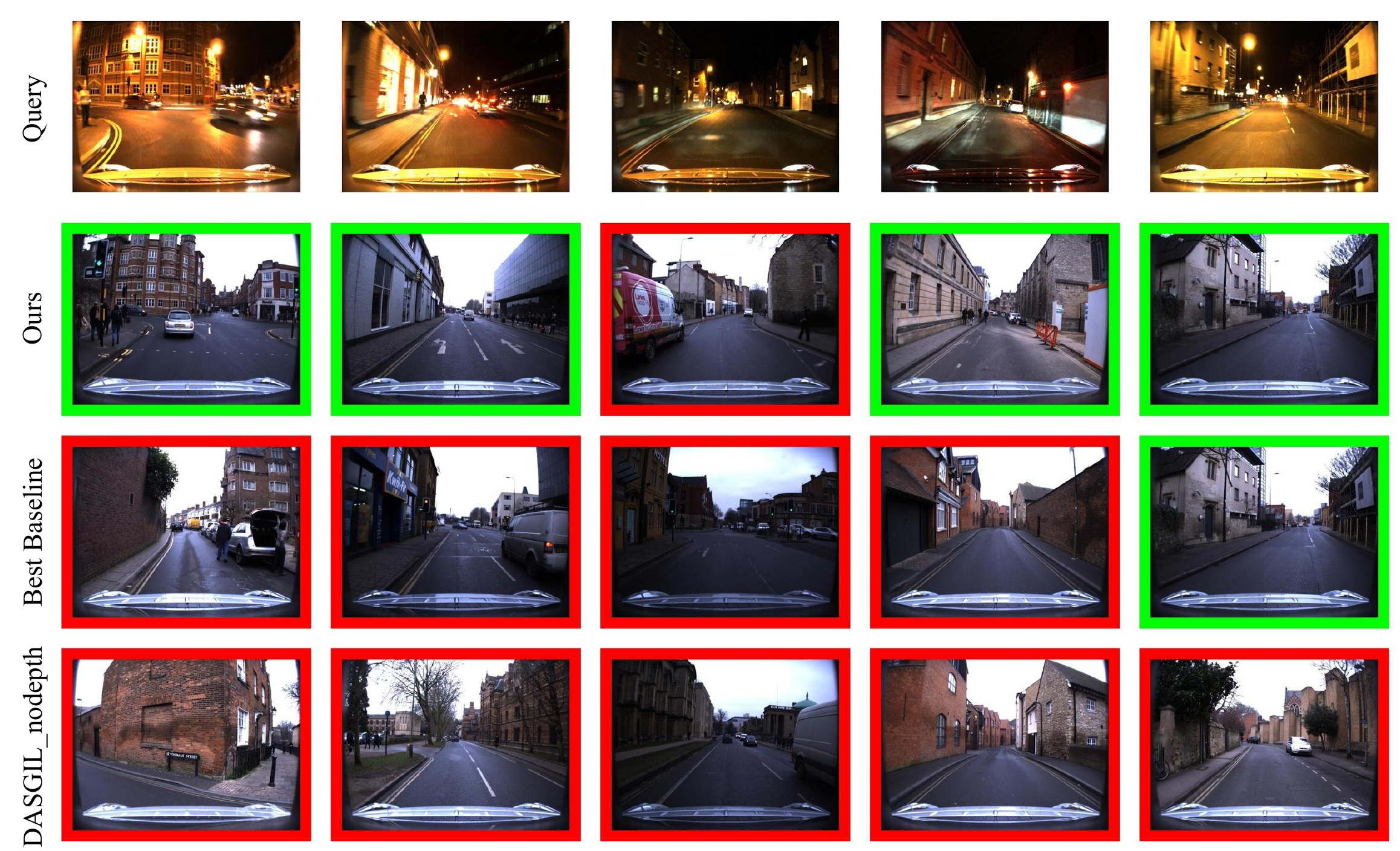}
        \caption{}
    \end{subfigure}
    \vspace{6pt}
    \hrule
    \vspace{6pt}
    \begin{subfigure}{1.0\textwidth}
        \includegraphics[width=\linewidth]{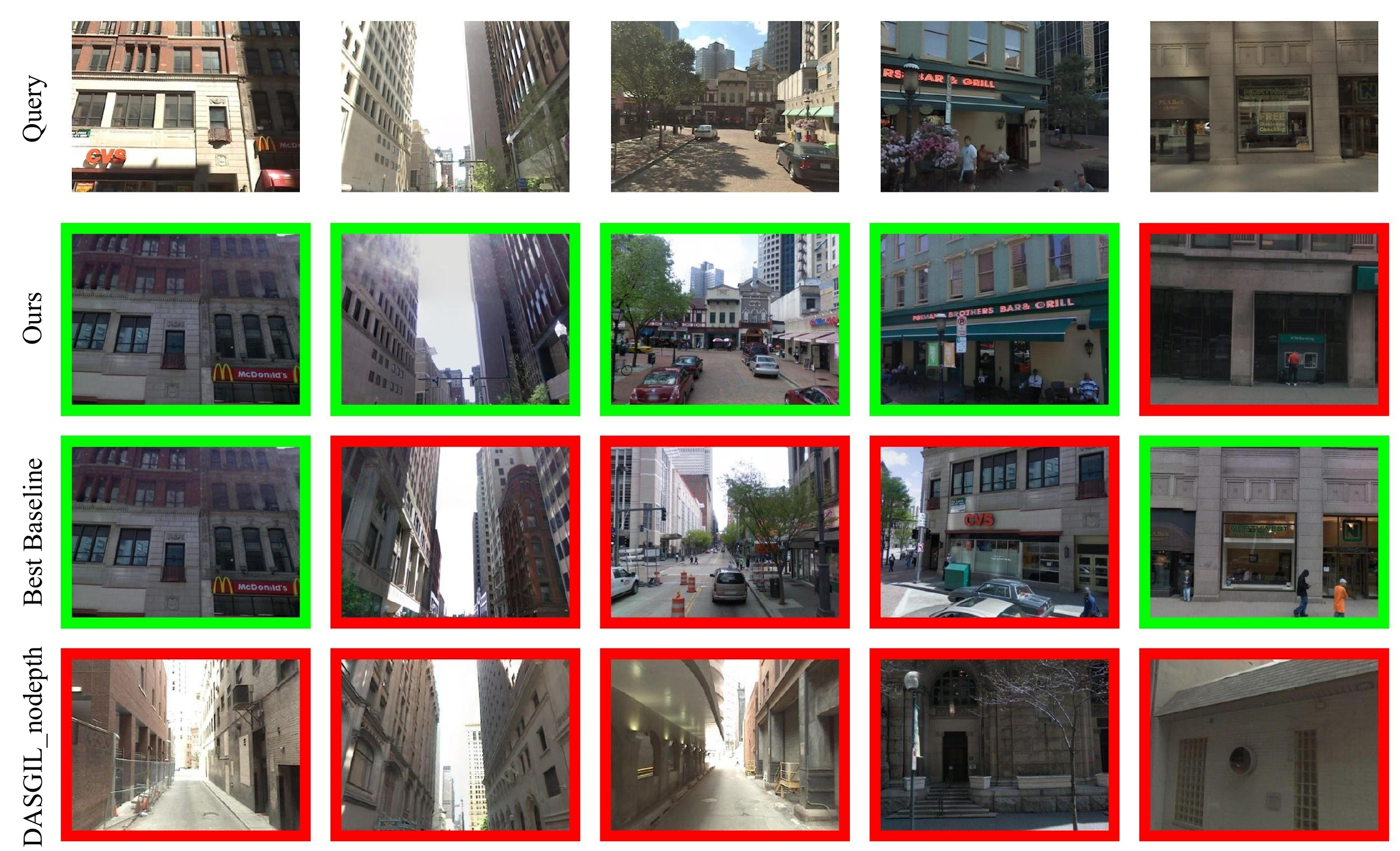}
        \caption{}
    \end{subfigure}
    \caption{Qualitative predictions testing on: a) Oxford RobotCar \cite{Maddern_IJRR2017} with queries from Night scenario and gallery from Overcast; b) Pitts30k \cite{Torii_2015PAMI}.}
    \label{fig:preds_night_and_pitts}
    \vspace{10pt}
\end{figure*}

\begin{figure*}[t!]
    \centering
    \includegraphics[width=1.0\linewidth]{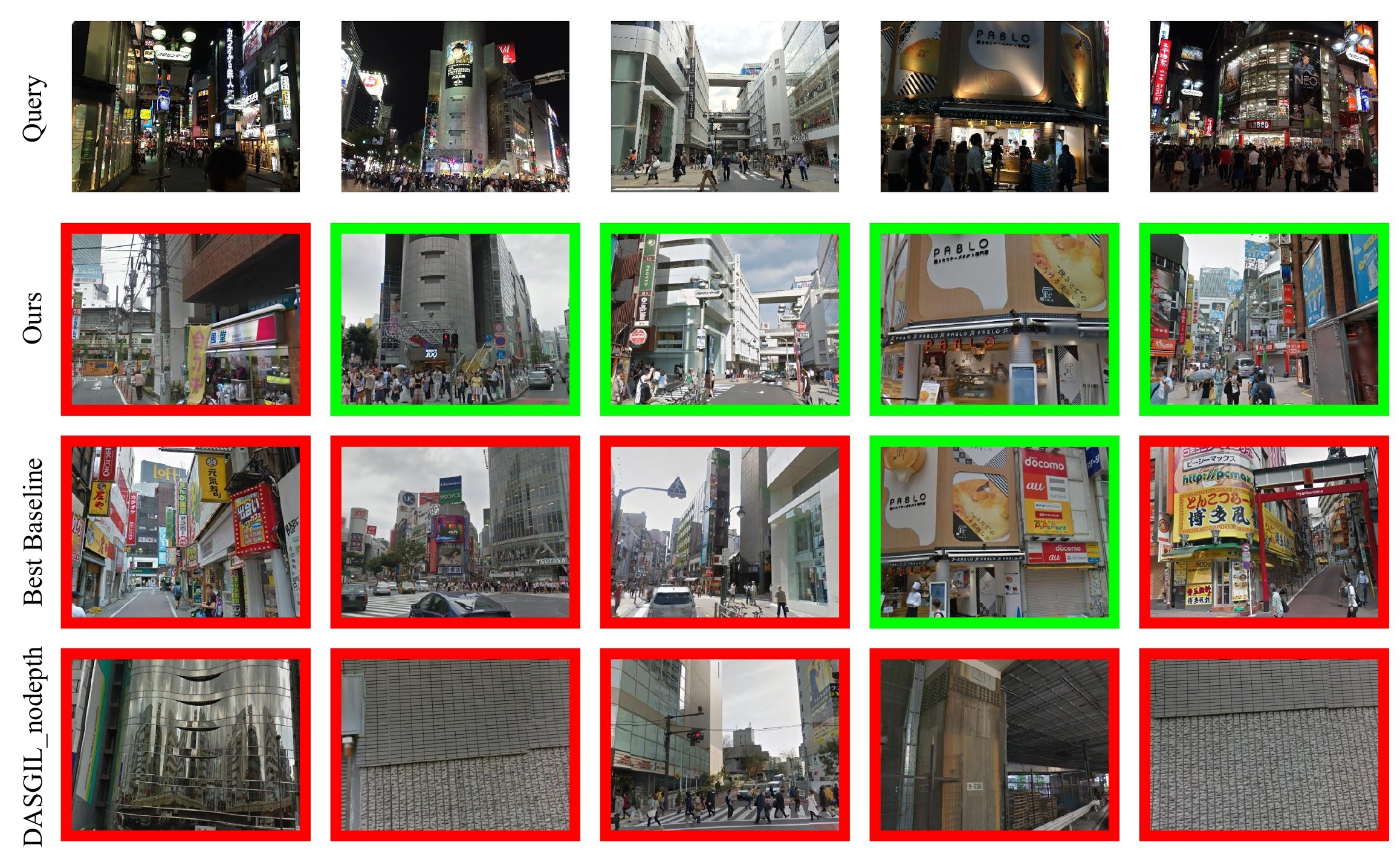}
    \caption{Qualitative predictions testing on RTokyo \cite{Warburg_2020_CVPR}}
    \label{fig:preds_rtokyo}
\end{figure*}

\noindent
\cref{fig:preds_rain}, \cref{fig:preds_snow_and_sun}, \cref{fig:preds_night_and_pitts}, \cref{fig:preds_rtokyo} provide a qualitative comparison between our framework and the best competitors tested in the companion paper: the best baseline (GeM \cite{Radenovic_PAMI2019} + DA $\star$) and DASGIL\_nodepth \cite{Hanjiang_TIP2021}.   
We use \emph{Ours w PSP $\star$} configuration, testing on Oxford RobotCar \cite{Maddern_IJRR2017} scenarios, Pitts30k \cite{Tolias_ICLR2016} and RTokyo \cite{Warburg_2020_CVPR} (as in the Sec. 5, 6 of the main paper).
All of the provided results confirm the assumption that by conditioning the semantic segmentation on the place recognition task our model learns to draw information only from the semantic categories that are most discriminative for a place, thereby outperforming the current state of the art on all the tested scenarios.   

\subsection{Extra experiments}
\noindent
In this section, we propose additional analysis comparing our method to NetVLAD \cite{Arandjelovic_CVPR2016}, a commonly used baseline in Visual Place Recognition (VPR), as well as other experiments involving our architecture and baselines trained on real-world datasets.

\paragraph{Comparisons with NetVLAD.}
\begin{table*}[t]
\begin{adjustbox}{width=1.0\textwidth}
\centering
\begin{tabular}{rcc||cc|cc}
\hline
  \begin{tabular}[c]{@{}c@{}}Method\end{tabular} &
  \begin{tabular}[c]{@{}c@{}}Oxford RobotCar Avg\\ 1 / 5 / 10\end{tabular} &
  \begin{tabular}[c]{@{}c@{}}Avg\end{tabular} &
  \begin{tabular}[c]{@{}c@{}}RTokyo - Test\\ 1 / 5 / 10\end{tabular} & 
  \begin{tabular}[c]{@{}c@{}}Avg\end{tabular} &
  \begin{tabular}[c]{@{}c@{}}MapillarySLS\\ 1 / 5 / 10\end{tabular} &
  \begin{tabular}[c]{@{}c@{}}Avg\end{tabular} \\
\hline

NetVLAD \cite{Arandjelovic_CVPR2016}              & 43.6 / 58.5 / 65.7 & 55.9 & 19.2 / 38.7 / 46.8 & 34.9 & 14.9 / 24.9 / 29.2 & 23.0 \\
NetVLAD \cite{Arandjelovic_CVPR2016} + DA         & 50.5 / 63.9 / 73.5 & 62.6 & 20.5 / 40.1 / 47.8 & 36.1 & 11.8 / 20.8 / 25.6 & 19.4 \\
NetVLAD \cite{Arandjelovic_CVPR2016} $\star$      & 45.7 / 61.3 / 68.9 & 58.6 & 16.8 / 34.7 / 38.7 & 30.1 & 23.1 / 35.3 / 41.5 & 33.3 \\
NetVLAD \cite{Arandjelovic_CVPR2016} + DA $\star$ & 53.3 / 67.3 / 75.5 & 65.3 & 23.6 / 41.4 / \textbf{52.9}  & 39.3 & 32.6 / 46.4 / 51.8 & 43.6 \\

\hline
Ours w DL \cite{Chen2018DeepLabSI}                & \textbf{58.5} / 71.1 / 76.5 & \textbf{68.7} & 20.7 / 37.1 / 45.5 & 34.4 & \textbf{34.6} / 45.8 / 52.3 & 44.2 \\
Ours w DL \cite{Chen2018DeepLabSI} $\star$        & 54.7 / 68.9 / 74.8 & 66.1 & 21.4 / 38.0 / 45.0 & 34.8 & 34.5 / 46.8 / \textbf{53.0} & \textbf{44.8} \\
Ours w PSP \cite{Zhao_CVPR2017}                   & 52.3 / 65.9 / 71.6 & 63.3 & 17.4 / 30.9 / 38.6 & 29.0 & 33.0 / \textbf{47.2} / \textbf{53.0} & 44.4 \\
Ours w PSP \cite{Zhao_CVPR2017} $\star$           & 55.6 / \textbf{71.5 / 77.4} & 68.2 & \textbf{26.9} / \textbf{44.6} / 51.9 & \textbf{41.1} & 32.9 / 46.1 / 51.9 & 43.6 \\ 
\hline
\end{tabular}
\end{adjustbox}
\vspace{2pt}

\caption{Additional experiments involving our novel synthetic dataset and the Oxford RobotCar scenarios \cite{Maddern_IJRR2017} (used for the DA branch). All of the experiments employ a ResNet50 as the backbone, with the exception of $\star$, which uses a ResNet101. DA stands for the domain adaptation branch. DeepLab V2 \cite{Chen2018DeepLabSI} is abbreviated as DL and PSPNet is abbreviated as PSP \cite{Zhao_CVPR2017}. The results to the left of the double line indicate the average recall over the RobotCar scenarios (Rain, Snow, Sun, and Night as the query set, whereas Overcast as the gallery),
while the results to the right of the double line show the models' ability to generalize to two previously unseen datasets, RTokyo \cite{Warburg_2020_CVPR} and MapillarySLS \cite{Warburg_2020_CVPR}, respectively.}
\label{tab:results_netvlad}
\vspace{-2pt}
\end{table*}

\noindent
We extended to NetVLAD \cite{Arandjelovic_CVPR2016} the experimental setup described in paragraphs 5.1 and 5.2 of the companion paper. As encoders, we used ResNet50 and ResNet101, and 64 cluster centroids inside the NetVLAD pooling layer, then trained the model using our newly synthetic dataset. Similarly to what we did for GeM \cite{Radenovic_PAMI2019} and RMAC \cite{Tolias_ICLR2016}, we replicated the configurations of \cite{Arandjelovic_CVPR2016}, adding unsupervised domain adaptation to reduce the shift between real and synthetic data. For this purpose, we leverage all the scenarios of the Oxford RobotCar dataset \cite{Maddern_IJRR2017}: Rain, Snow, Sun and Night. 

\cref{tab:results_netvlad} presents two experimental groups and shows the 1/5/10 recall as well as the overall average accuracy. The first set of experiments (to the left of the double line) refers to the validation on RobotCar \cite{Maddern_IJRR2017}, which adheres to the same protocol as section 5 of the main paper. The Overcast scenario serves as the gallery, and the Rain, Snow, Sun, and Night scenarios serve as the query set. The results demonstrate how our strategy outperforms the baseline by $+3.4\%$.
The second set of experiments (to the right of the double line) is designed to demonstrate the generalization capability of our model over images never seen during training. We evaluate the effectiveness of our approach on the two RTokyo (a revisited version of Tokyo24/7 \cite{Torii_CVPR2015} proposed by \cite{Warburg_2020_CVPR}) and MapillarySLS \cite{Warburg_2020_CVPR} datasets. Also in this case, the results demonstrate the usefulness of our strategy in generalizing to a whole new target, yielding a boost of $+1.8\%$ and $+1.2\%$ to the two evaluated datasets, respectively.
Note that the architecture proposed by \cite{Arandjelovic_CVPR2016} produces $131072$-D (dimensional) descriptors \wrt the more compact (around 43 times lower) $3072$-D descriptors produced by our method. Descriptors with these proportions representing large datasets, such as RTokyo or MapillarySLS, not only require a large amount of RAM to be saved concurrently, but also increase the time required to complete the retrieval (k-NN) between query and gallery. \cref{tab:time_consumption} confirms this by reporting the expected inference time (descriptor extraction and k-NN) using the same resources (NVIDIA TITAN X GPU and Intel Core i7-5930K CPU @ 3.50GHz CPU) and full resolution dataset.
\begin{table}[t]
\centering
\begin{adjustbox}{width=0.9\linewidth}
\begin{tabular}{rccc}
\hline
\begin{tabular}[c]{@{}c@{}}Method\end{tabular} &
\begin{tabular}[c]{@{}c@{}}Descriptor \\ dimension\end{tabular} &
\begin{tabular}[c]{@{}c@{}}RTokyo\\64608 Gallery Imgs\\247 Query Imgs\end{tabular} &
\begin{tabular}[c]{@{}c@{}}MapillarySLS\\18920 Gallery Imgs\\11120 Query Imgs\end{tabular} \\
\hline
NetVLAD \cite{Arandjelovic_CVPR2016} & 131072-D & 1:25:19 & 0:38:53 \\
Ours                                 & 3072-D   & 0:35:13 & 0:16:43 \\
\hline
\end{tabular}
\end{adjustbox}
\vspace{2pt}
\caption{Comparison of the inference time \big[h:mm:ss\big] between NetVLAD \cite{Arandjelovic_CVPR2016} and our technique over the RTokyo \cite{Warburg_2020_CVPR} and MapillarySLS \cite{Warburg_2020_CVPR} test sets.}
\label{tab:time_consumption}
\end{table}

\begin{table}[t]
\centering
\begin{adjustbox}{width=0.7\linewidth}
\begin{tabular}{rcc}
\hline
\begin{tabular}[c]{@{}c@{}}Method\end{tabular} &
\begin{tabular}[c]{@{}c@{}}Pitts30k - Val\\ 1 / 5 / 10\end{tabular} &
\begin{tabular}[c]{@{}c@{}}Pitts30k - Test\\ 1 / 5 / 10\end{tabular} \\
\hline
GeM \cite{Radenovic_PAMI2019}        & 75.2 / 90.5 / 93.9 & 73.3 / 86.9 / 90.7 \\
RMAC \cite{Tolias_ICLR2016}          & 73.5 / 90.5 / 94.6 & 74.1 / 88.5 / 92.7 \\
NetVLAD \cite{Arandjelovic_CVPR2016} & 79.5 / \textbf{93.0} / \textbf{96.1} & 77.3 / 88.8 / 92.1 \\
\hline
Ours w DL \cite{Chen2018DeepLabSI}   & \textbf{81.9} / 92.7 / 95.8 & \textbf{79.8} / \textbf{90.7} / \textbf{93.5} \\
\hline
\end{tabular}
\end{adjustbox}
\vspace{2pt}
\caption{Experiments using Pitts30k \cite{Torii_2015PAMI} training set previously segmented with PSPNet \cite{Zhao_CVPR2017} pretrained on ADE20k \cite{Zhou_CVPR2017_ADE20k}. Evaluation results using Pitts30k \cite{Torii_2015PAMI} validation and test sets.}
\label{tab:results_real_world_setting}
\end{table}

\paragraph{Real-world dataset with imperfect semantic labels.}
\noindent
Despite the usage of a domain adaptation module, the results with all the models trained on the synthetic dataset suffer from the considerable gap with the real-word target domain.
To demonstrate that our solution provides a performance uplift against the comparison even when all the methods are trained on real-world data, we conducted a series of experiments on a real-world VPR dataset coarsely annotated using a pretrained semantic segmentation model. More specifically, we annotated the Pitts30k \cite{Torii_2015PAMI} train set using a PSPNet \cite{Zhao_CVPR2017} model, with a ResNet50 as a backbone, pretrained on ADE20k \cite{Zhou_CVPR2017_ADE20k}.
Note that this is a disadvantageous scenario for our method because of the imperfect semantic labels, which limit how well the model can learn to exploit the semantic content of the scene.
\cref{tab:results_real_world_setting} presents the results of GeM \cite{Radenovic_PAMI2019}, RMAC \cite{Tolias_ICLR2016}, NetVLAD \cite{Arandjelovic_CVPR2016}, and our novel approach. 
Nonetheless, our work remains cutting-edge, demonstrating that its usage of the semantic information guided by the place recognition task holds promise to achieve even better results when a real-world dataset with precise pixel-wise semantic labels becomes available.

%
%
%
%
%
%
%
%
%
%
{\small
\newpage
\bibliographystyle{splncs04}
\bibliography{egbib}
}
\end{document}